\documentclass[10pt,twocolumn,letterpaper]{article}

\usepackage[table]{xcolor}

\usepackage{cvpr}
\usepackage{times}
\usepackage{epsfig}
\usepackage{graphicx}
\usepackage{amsmath}
\usepackage{amssymb}

\usepackage{subfigure}
\usepackage{dsfont}
\usepackage{tabularx}
\usepackage{booktabs}
\usepackage{tabu}
\usepackage{makecell}

\usepackage{color}

\newcommand{\comment}[1]{}

\newcommand{\note}[1]{\noindent\framebox{\begin{minipage}{8cm}{\bf #1}\end{minipage}}\\}

\newcommand{\bof}{\mathbf{f}}

\newcommand{\bW}{\mathbf{W}}

\newcommand{\by}[0]{\mathbf{y}}

\DeclareMathOperator*{\argmax}{arg\,max}

\newcommand{\fig}[1]{Fig.~\ref{fig:#1}}

\newcommand{\tbl}[1]{Table~\ref{tbl:#1}}

\definecolor{orange}{rgb}{1,0.5,0}
\definecolor{blue}{rgb}{0,0,0.6}

\definecolor{color1}{RGB}{0,199,1}
\definecolor{color2}{RGB}{224,43,28}

\usepackage[pagebackref=true,breaklinks=true,letterpaper=true,colorlinks,bookmarks=false]{hyperref}

\usepackage{snapshot}           

\cvprfinalcopy 

\def\cvprPaperID{141} 

\ifcvprfinal\fi
\begin{document}


\title{Learning to Assign Orientations to Feature Points}


\author{Kwang Moo Yi\textsuperscript{1,}\thanks{First two authors
    contributed equally.} \qquad  Yannick Verdie \textsuperscript{1,}\footnotemark[1]  \qquad  Pascal Fua\textsuperscript{1} \qquad    Vincent Lepetit\textsuperscript{2}\\
  \textsuperscript{1}Computer Vision Laboratory, \'{E}cole Polytechnique F\'{e}d\'{e}rale de Lausanne (EPFL)\\
  \textsuperscript{2}Institute for Computer Graphics and Vision, Graz University of Technology\\
  {\tt\small    \{kwang.yi,   yannick.verdie,    pascal.fua\}@epfl.ch,
    {\tt\small lepetit@icg.tugraz.at}} }
\maketitle


\begin{abstract}

  We show  how to  train a  Convolutional Neural Network  to assign  a canonical
  orientation to  feature points given  an image  patch centered on  the feature
  point.  Our method  improves feature point matching upon  the state-of-the art
  and  can  be  used  in   conjunction  with  any  existing  rotation  sensitive
  descriptors.  To  avoid the tedious  and almost  impossible task of  finding a
  target  orientation  to  learn,  we  propose to  use  Siamese  networks  which
  implicitly find  the optimal orientations  during training. We also  propose a
  new  type of  activation function  for  Neural Networks  that generalizes  the
  popular ReLU, maxout,  and PReLU activation functions.   This novel activation
  performs better  for our task.   We validate  the effectiveness of  our method
  extensively with four existing datasets, including two non-planar datasets, as
  well as  our own  dataset.  We  show that  we outperform  the state-of-the-art
  without the need of retraining for each dataset.

\end{abstract}



\section{Introduction}


Feature points  are an  essential and  ubiquitous tool  in computer  vision, and
extensive research  has been conducted on  both detectors~\cite{Alcantarilla12b,
  Bay06,    Leutenegger11,   Lowe04,    Miko04c,    Rublee11,   Zitnick11}    and
descriptors~\cite{Alahi12,  Bay06,  Leutenegger11,   Lowe04,  Rublee11,  Tola10,
  Winder09,        Zitnick10},         including        using        statistical
approaches~\cite{Simo-Serra15,  Zagoruyko15}.   However,  the  assignment  of  a
canonical orientation, which is an important common step, has received almost no
individual  attention,   probably  since   the  dominant  orientation   of  {\it
  SIFT}~\cite{Lowe04} is considered to give good results.

\def \teaserWidth {0.205}
\addtocounter{footnote}{-1}\addtocounter{Hfootnote}{-1}
\begin{figure} \centering
  \subfigure{
    \includegraphics[width = \teaserWidth\textwidth,trim = 8 8 8 8,clip]{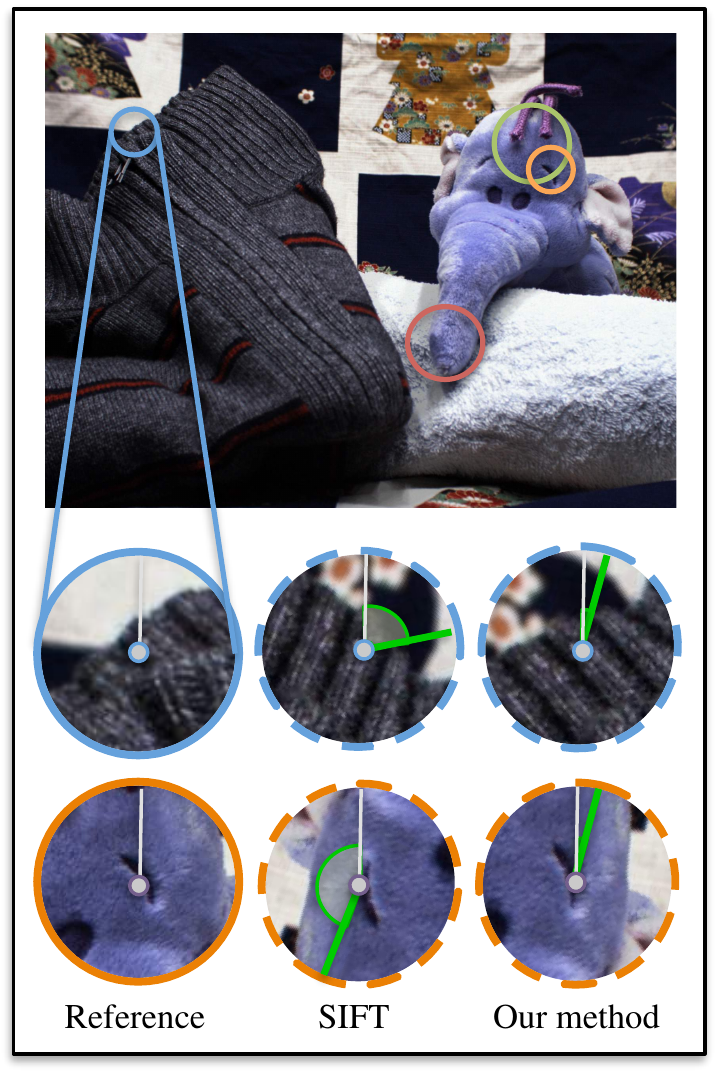}
  }
\hspace{-0.75em}
  \subfigure{
    \includegraphics[width = \teaserWidth\textwidth,trim = 8 8 8 8,clip]{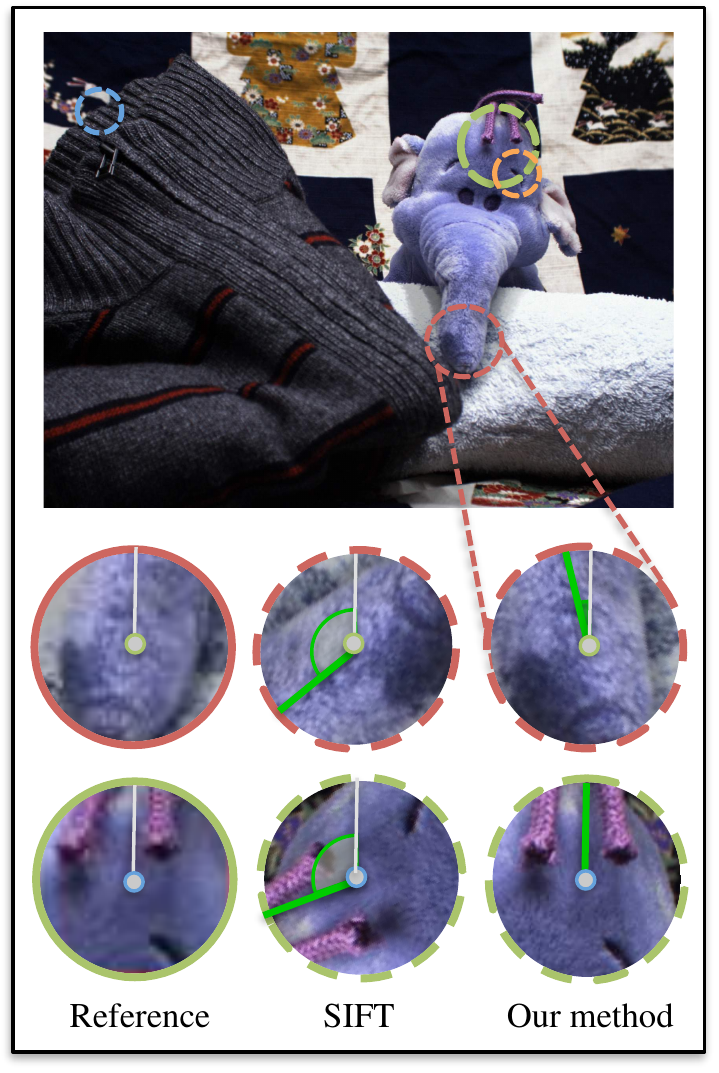}
  }
\vspace{-1.0em}

  \subfigure{
    \includegraphics[width = \teaserWidth\textwidth, trim=18 20 5 20,clip]{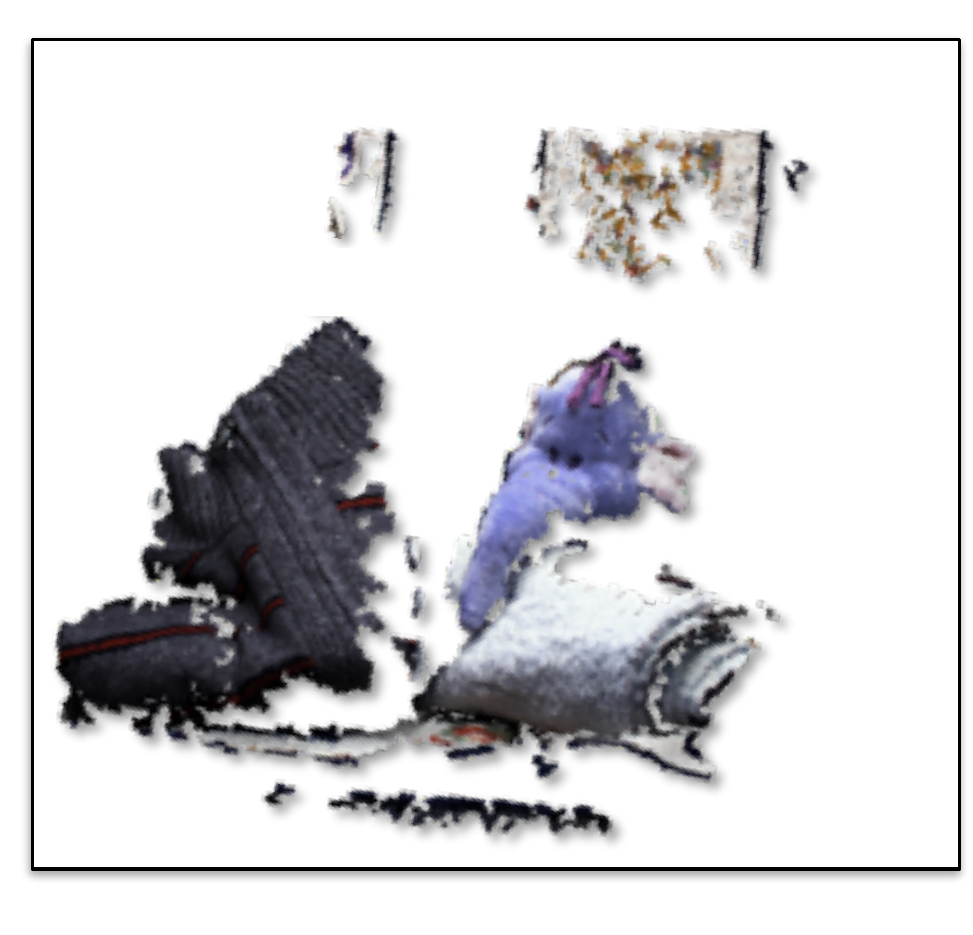}
  }
\hspace{-1.1em}
  \subfigure{
    \includegraphics[width = \teaserWidth\textwidth, trim=18 18 15 10,clip]{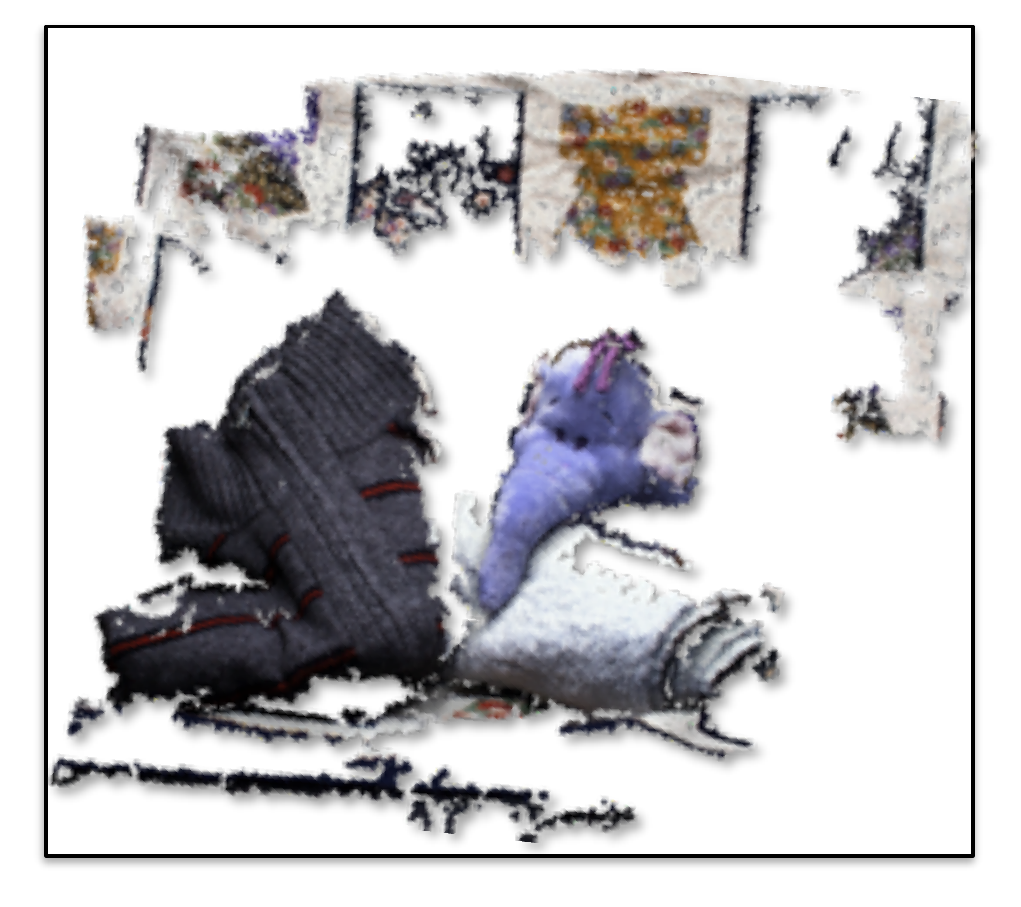}
  }

  \caption{Multi-View  Stereo  (MVS)~\cite{Wu13,Wu11}  reconstruction  with  the
    orientation  assignment  of  {\it SIFT}~\cite{Lowe04}  and  our  orientation
    assignment.  \textbf{Top:}  two of the original  images from~\cite{Aanaes12}
    used for MVS.  \textbf{Middle:} enlarged  feature point regions in groups of
    three; {\it left} reference region from  the left image, {\it center} region
    rotated back with  {\it SIFT} orientations, {\it right}  region rotated back
    with our learned orientations. Estimation errors are denoted by a green arc.
    \textbf{Bottom  left:}  MVS  results   with  {\it  SIFT}  orientations,  and
    \textbf{Bottom right:} MVS results with  our orientations.  As shown, due to
    viewpoints changes on  non-planar surfaces, {\it SIFT}  orientations are not
    stable.  On  the contrary, orientations  provided by our method  are stable,
    which leads to  better reconstructions.  46272 vertices  were obtained using
    {\it SIFT}  orientations, and  84087 vertices  with our  orientations.  {\it
      Edge Foci}  feature points~\cite{Zitnick11} were used  in conjunction with
    {\it     Daisy}~\cite{Tola10}     descriptors      for     both     methods.
    \protect\footnotemark}
  \label{fig:teaser}
\end{figure}
\footnotetext{Figures are best viewed in color.}

%


However, this is not necessarily true.  In complex scenes, feature points lie on
non-planar surfaces and their appearance can be drastically altered by viewpoint
and  illumination  changes.   This  can easily  produce  errors  in  orientation
estimates   as  shown   in  \fig{teaser}.    In  addition,   rotation  invariant
descriptors~\cite{Bellavia10,  Fan11, Wang11e}  are  not  a definitive  solution
either as these descriptors discard  rotation sensitive information which can be
useful  when  ideal  orientations are  given.  Thus,  as  we  will show  in  our
experiments,  higher  matching  performances   can  be  achieved  with  rotation
sensitive descriptors and better orientation assignments.




In this  paper, we show how  to remedy this  problem by training a  regressor to
estimate  better orientations  for matching,  and  to boost  the performance  of
existing rotation sensitive descriptors. We train a Convolutional Neural Network
to predict an orientation,  given a patch around a feature  point.  To avoid the
difficult  task of  finding the  canonical orientation  to learn,  we treat  the
orientation  to   learn  as  an   implicit  variable,  by  training   a  Siamese
network~\cite{Bromley93,Chopra05}     similar     to     descriptor     learning
methods~\cite{Simo-Serra15, Zagoruyko15}.  Also, to allow  our method to work in
conjunction  with  any existing  rotation  sensitive  descriptors such  as  {\it
  SIFT}~\cite{Lowe04},  {\it  SURF}~\cite{Bay06},  and the  learning-based  {\it
  VGG}~\cite{Simonyan14}, we  consider the descriptor  component as a  black box
when learning.

  


We also propose  a new activation function for the  neurons based on Generalized
Hinging Hyperplanes~(GHH)~\cite{Wang05d}, which plays a  key role in our method.
We will show that it generalizes the popular ReLU and maxout~\cite{Goodfellow13}
activation  functions,  as  well  as  the  recent  PReLU~\cite{He15}  activation
function, with better performance for our task.

To evaluate the performance of  descriptors with orientations from the proposed
method,    we    use    datasets    with    both    planar    or    far    away
objects~\cite{Miko04b,Verdie15,   Zitnick11}  and   3D  objects~\cite{Aanaes12,
  Strecha08b}.  In  addition, we created  our own  dataset as well,  to further
enrich the  dataset with complex  camera movements, such as  in-plane rotations
and  viewpoint  changes.   We  demonstrate   that  the  proposed  method  gives
significant improvement over the  state-of-the-art for \emph{all} the datasets,
without the need of re-training for each dataset.

In the  remainder of this  paper, we first  discuss related work,  introduce our
learning  framework,  detail our  method  as  well  as the  proposed  activation
function.   We   then  present   our  experimental  results   demonstrating  the
effectiveness of our orientation assignment compared to the state-of-the-art. We
also investigate the  influence of the proposed activation and  of the datasets,
and we conclude with several application results.

\section{Related Work}

As shown  in the survey  of~\cite{Gauglitz11}, the  importance of having  a good
orientation  estimation has  been  overlooked,  and thought  to  be  a not  very
important step which either feature point detector or descriptor has to perform.
The widely-used solution  for assigning an orientation to a  feature point is to
use the dominant  orientation of {\it SIFT}~\cite{Lowe04}.   However, as pointed
out by~\cite{Liu14},  dominant orientation-based  methods do  not work  well for
arbitrary  positions,  although  it  has   critical  impact  on  the  descriptor
performances~\cite{Lin12}.   Nevertheless, here  we  provide a  brief review  of
existing methods related to orientation assignment and our method.

\vspace{-0.8em}
\paragraph{Orientation assignment of feature point detectors.}

In {\it  SIFT}~\cite{Lowe04}, histograms  of gradient  orientations are  used to
determine the dominant orientation.  It remains  the most popular method and has
also  been  extended  to   3D~\cite{Allaire08}.   {\it  SURF}~\cite{Bay06}  uses
Haar-wavelet   responses   of   sample    points   to   extract   the   dominant
orientation. {\it MOPs}~\cite{Brown05a}  simply uses the gradient  at the center
of   a   patch  after   some   smoothing   for   robustness  to   noise.    {\it
  ORB}~\cite{Rublee11} uses image moments to compute  the center of mass as well
as  the   main  orientation.   {\it  HIP}~\cite{Taylor11}   considers  intensity
differences over  a circle  centered around  the feature  point to  estimate the
orientation. Although this is rather fast, it is also very sensitive to noise.

In summary,  despite the variation,  the main idea  of these methods  remain the
same: finding  a reliable dominant  orientation in their respective  ways. Thus,
when  computation  time  constraints  are   not  too  drastic,  using  the  SIFT
orientation remains to be the first solution to try~\cite{Gauglitz11}.

\vspace{-0.8em}
\paragraph{Rotation invariant descriptors.}

As  existing orientation  assignment methods  are  not always  robust enough  to
guarantee good  matching performances,  interest has  been drawn  to descriptors
which   are   inherently   rotation   invariant~\cite{Fan11,   Wang11e}.    {\it
  MROGH}~\cite{Fan11} uses local intensity order pooling with rotation invariant
gradients, and {\it LIOP}~\cite{Wang11e} constructs  the descriptor in a similar
way  but with  a different  strategy for  aggregating the  gradient information.
{\it BRISK}~\cite{Leutenegger11} and  {\it FREAK}~\cite{Alahi12} claims rotation
invariance as well, but they still depend on the orientation estimation which is
included in the descriptor extraction process.

Besides   descriptors    that   are   rotation   invariant    by   construction,
\cite{Lazebnik04}  uses concentric  rings for  generating orientation  histogram
bins with spin  images, and a specific distance function  for rotation invariant
matching.   {\it  sGLOH}~\cite{Bellavia10}  also  proposes  to  use  a  rotation
invariant distance function  which computes distances for  all possible rotation
combinations and takes the minimum.  The  authors further extend their method by
proposing a general  method for histogram-based feature  descriptors taking into
account the main orientation of the scene~\cite{Bellavia14}.

Although   these  methods   may  be   better  than   the  original   {\it  SIFT}
descriptor~\cite{Lowe04}, {\it SIFT} descriptor combined with our learning-based
orientation estimation  outperforms them,  as we will  show in  the experiments.
This  is  probably due  to  the  fact  that  rotation sensitive  information  is
discarded when  computing these descriptors.  Furthermore,  \cite{Bellavia14} is
only  applicable  when the  entire  scene  is the  object  of  interest, and  is
impractical as  the main orientation is  obtained by computing all  the possible
matching pairs of features to keep the configuration with best matches.

\vspace{-0.8em}
\paragraph{Learning-based methods.}
Learning-based methods  have been already used  in the context of  feature point
matching, but  only for  problems other than  orientation assignment  of general
feature points.  For example,  \cite{Hinterstoisser08a} learns to
predict the pose of patches, but uses one regressor per patch, which
is not a viable solution for general feature points.
\cite{Simo-Serra15, Zagoruyko15} use Siamese  networks---as we do---to directly
compare   image   patches~\cite{Zagoruyko15},   or    to   learn   to   compute
descriptors~\cite{Simo-Serra15}.   {\it   VGG}~\cite{Simonyan14}  as   well  as
\cite{Fan14,Trzcinski15} also  learn descriptors, through  convex optimization,
boosting, and greedy optimization, respectively.

One caveat  in these learning-based descriptors  is that they still  rely on the
orientation  estimation  of  local  feature detectors  which  are  traditionally
handcrafted.  Moreover, they typically  use the Brown dataset~\cite{Brown10} for
learning, with patches extracted using  ground truth orientations from Structure
from Motion~(SfM) techniques.   This ground truth orientation  assignment is not
something one can expect  to have in practical use, and  may lead to performance
degradation   when   tested   on   other  data   with   inaccurate   orientation
assignments~\cite{Simonyan14}.   These methods  will  also  benefit from  better
orientation assignments  on test time, as  we will show in  our experiments with
{\it VGG}.



\section{Method}

In this section we first introduce our learning strategy, then formalize it.  We
also describe our activation function based on GHH.

\subsection{Canonical Orientation as an Implicit Variable}

\begin{figure}
  \centering
  \subfigure[Matches with {\it SIFT} orientations]{
    \includegraphics[width = 0.22\textwidth, trim=40 40 40 25, clip]{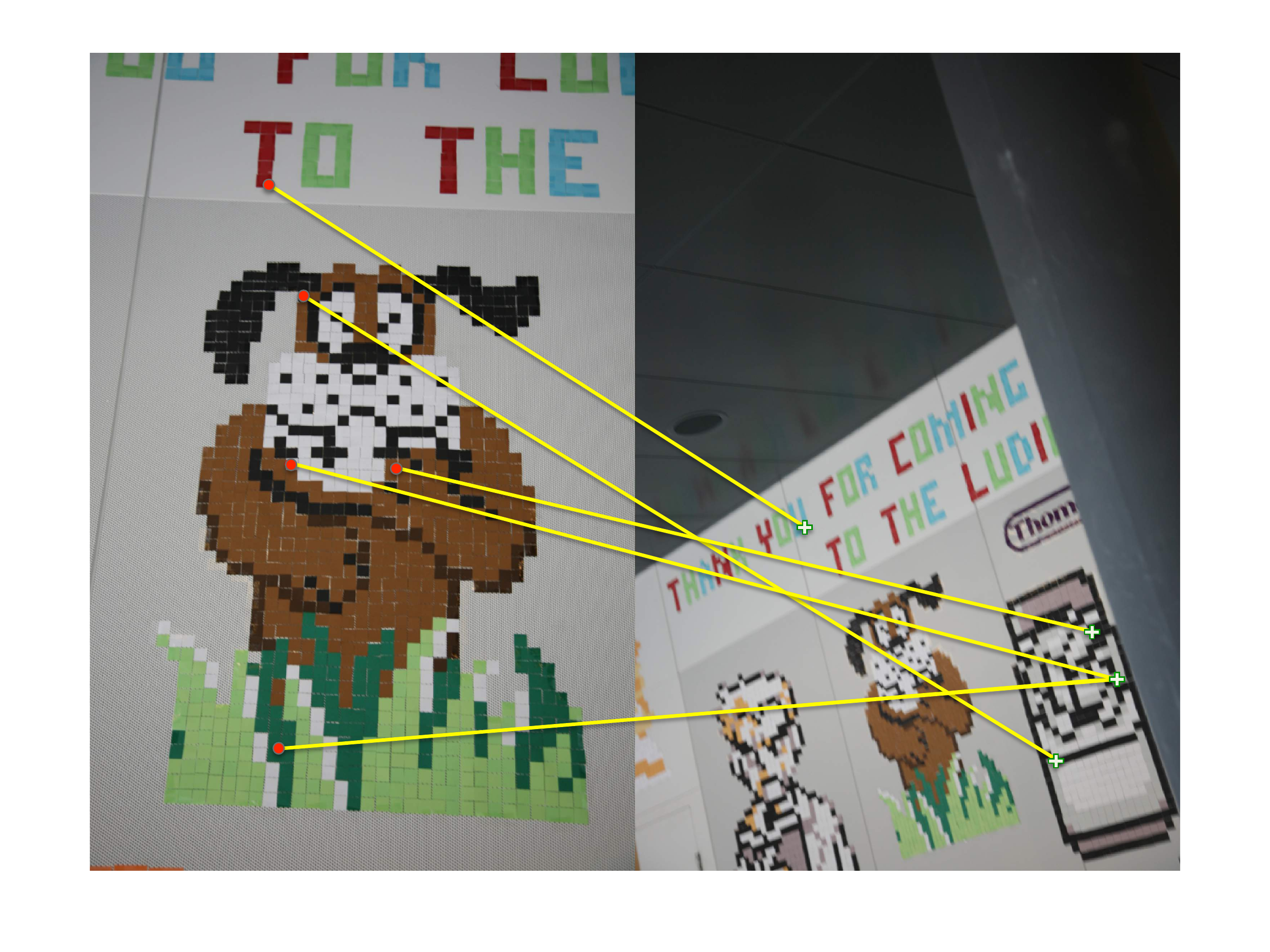}
  }
  \subfigure[Matches with our orientations]{
    \includegraphics[width = 0.22\textwidth, trim=40 40 40 25, clip]{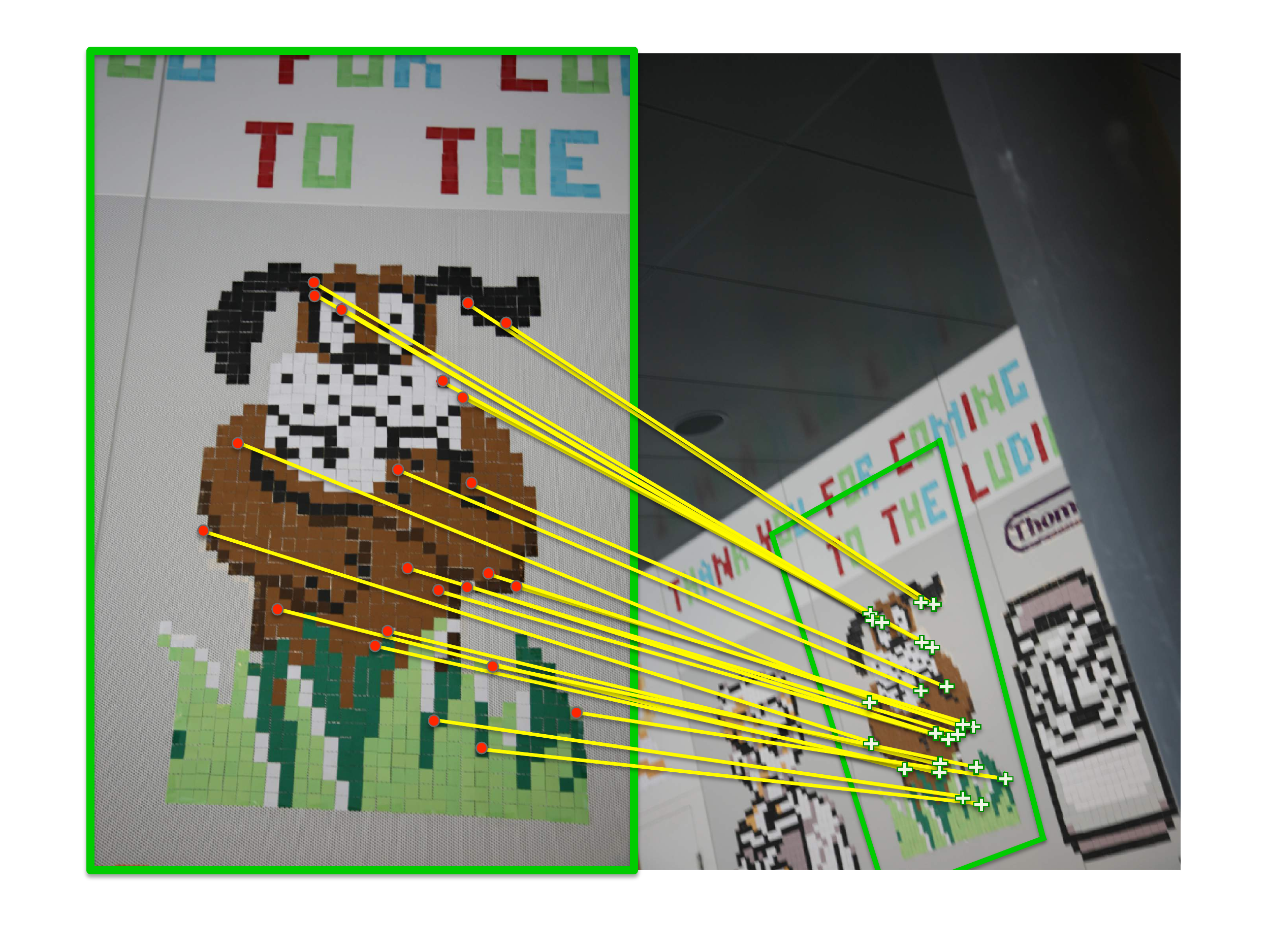}
  }

  \caption{Image  matching  example  from  the duckhunt  sequence  in  the  {\it
      Viewpoints}  dataset with  {\it  SIFT} descriptors  using the  orientation
    assigned by (a) {\it SIFT} and (b)  our method.  The yellow lines denote the
    inlier matches  after RANSAC.   The homography  is correctly  estimated only
    when using our orientations.  }
  \label{fig:orientation}
\end{figure}

As  illustrated in  \fig{orientation}, orientation  assignment plays  a critical
role in the descriptor matching performances.   However, a major problem we face
in our approach is that it is not clear which orientation should be learned. For
example,  one can  try to  learn  to predict  the dominant  orientation of  {\it
  SIFT}~\cite{Lowe04}, or maybe the median  of the dominant orientations for the
same  feature  point extracted  from  multiple  images.   However, there  is  no
guarantee  that the  orientations  retrieved  from such  approach  is the  ideal
canonical orientation  we want to learn.   Our early experiments, based  on such
heuristics to decide which orientation should be learned, remained unfruitful.

Since it  is hard to  define a canonical orientation  to learn, we  instead take
into account that it is actually  the descriptor distances of the feature points
that are  important, not  the orientation values  themselves.  We  formulate the
problem  by  learning  to  assign orientations  which  minimize  the  descriptor
distances of pairs  of local features corresponding to the  same physical point.
In this way, we  do not have to decide which orientations  should be learned. We
let  the  learning  optimization  find  which  orientations  are  both  reliably
predictable and improve the matching  performance. We formalize this approach in
the next subsection.

\newcommand{\loss}{\mathcal{L}}
\newcommand{\bofi}[1]{\bof_i^{(#1)}}
\newcommand{\patch}{{\bf p}}
\newcommand{\rot}[1]{\text{rot}\Big(#1\Big)}
\newcommand{\desc}[1]{\text{desc}(#1)}

\subsection{Formalization}
\label{sec:formalization}

Our   approach   is   related   to  Siamese   networks   used   for   descriptor
learning~\cite{Simo-Serra15,  Zagoruyko15},  but  the   loss  function  and  its
computation are different since we learn to estimate the orientation and not the
descriptor itself.   In fact,  we treat the  descriptor as a  black box  so that
various  rotation  variant  descriptors  can  be used.   However,  this  is  not
necessarily a restriction, and can be  easily adapted to include learning of the
descriptors as well.

Our training data is  made of pairs of image patches  centered on feature points
in two images but  corresponding to the same physical 3D  points.  We minimize a
loss function $\sum_i \loss_i$  over the parameters $\bW$ of
a CNN, with
\begin{equation}
  \loss\left(\patch_i\right) = \left\|
  g(\patch_i^1, f_\bW(\patch_i^1)) -
  g(\patch_i^2, f_\bW(\patch_i^2))
  \right\|^2_2
  \;\;,
  \label{eq:loss}
\end{equation}
where          $\loss_i=\loss\left(\patch_i\right)$,          the          pairs
$\patch_i  = \{\patch_i^1,  \patch_i^2\}$ are  pairs of  image patches  from the
training  set, $f_\bW(\patch_i^*)$  denotes the  orientation computed  for image
patch    $\patch_i^*$    using    a    CNN   with    parameters    $\bW$,    and
$g(\patch_i^*,  \theta_i^*)$  is  the  descriptor  for  patch  $\patch_i^*$  and
orientation $\theta_i^*$.  As discussed in  the previous subsection, there is no
target orientation  in the loss  function of Eq.~\eqref{eq:loss}:  the predicted
orientations will be optimized implicitly during training.

\newcommand{\arctantwo}{\arctan\!2}

\vspace{-0.5em}
\paragraph{Predicting an angle.}

Learning angles requires a special care. Directly predicting an angle with a CNN
did not work well in our  early experiments, probably because the periodicity of
$f_\bW(\patch_i^*)$  in Eq.~\eqref{eq:loss}  generates  many  local minima.   An
alternative way would be to learn  to provide histogram-like outputs, which then
can be  used with  $\argmax$ to give  angular outputs, in  a way  reminiscent of
SIFT.   However,  this  approach  also  did not  work  well,  as  the  estimated
orientations have  to be discretized and  the network becomes too  large when we
want fine resolutions.

To alleviate the  problem of periodicity, similarly to how  manifolds are embed
in~\cite{Osadchy07,Penedones11}, we train a CNN $\hat{f}_\bW(.)$ to predict two
values, which can be seen as a scaled  cosine and sine, and compute an angle by
taking:
\begin{equation}
  f_\bW(\patch_i^*) = \arctantwo(\hat{f}_\bW^{(1)}(\patch_i^*), \hat{f}_\bW^{(2)}(\patch_i^*)) \;\; ,
\end{equation}
where  $\hat{f}_\bW^{(1)}(\patch_i^*)$  and $\hat{f}_\bW^{(2)}(\patch_i^*)$  are
the two values returned by the CNN for patch $\patch_i^*$, and $\arctantwo(y,x)$
is the  four-quadrant inverse  tangent function\footnote{We follow  the standard
  implementation for the C language for this function.}.

This function  is not defined at  the origin, which  turned out to be  a problem
only  happening  in  rare  occasions   at  the  first  iteration,  after  random
initialization  of the  CNN  parameters  $\bW$.  To  prevent  this,  we use  the
following approximation for its gradient:
\begin{equation}
  \nabla{\arctantwo}(y,x) = \left( 
    \frac{-y}{x^2+y^2+\epsilon},
    \frac{x}{x^2+y^2+\epsilon} 
  \right)
\;\;,
\label{eq:atangrad}
\end{equation}
where $\epsilon$ is a very small value.

\vspace{-0.5em}
\paragraph{Computing the derivatives.}

\newcommand{\bg}{{\bf g}}

The   derivatives   of  the   loss   function   for   a  given   training   pair
$\patch_i$ can be computed using the chain rule:
\begin{equation}
  \frac{\partial \loss_i}{\partial \bW} (\patch_i)
  = 
  \frac{\partial \loss_i}{\partial \bg_{1,i}} 
  \frac{\partial \bg_{1,i}}{\partial \theta_{1,i}} 
  \frac{\partial \theta_{1,i}}{\partial \bW}(\patch_i^1)
  +
  \frac{\partial \loss_i}{\partial \bg_{2,i}} 
  \frac{\partial \bg_{2,i}}{\partial \theta_{2,i}} 
  \frac{\partial \theta_{2,i}}{\partial \bW}(\patch_i^2)
  \;\;,
\label{eq:jacobian}
\end{equation}
with  $\theta_{*,i}   =  f_\bW(\patch_i^*)$   and  $\bg_{*,i}   =  g(\patch_i^*,
\theta_{*,i})$.

Jacobians      $\frac{\partial       \loss_i}{\partial      \bg_{*,i}}$      and
$\frac{\partial  \theta_{1,i}}{\partial \bW}$  are  straightforward to  compute.
$\frac{\partial  \bg_{*,i}}{\partial  \theta_{*,i}}$  is   not  as  easy,  since
$\bg_{*,i}$ is the descriptor for patch $\patch_i^*$ after rotation by an amount
given by  $\theta_{*,i}$. For  example, in  case of  {\it SIFT},  the descriptor
extraction process involves building histograms,  which cannot be expressed as a
differentiable function.  Moreover, depending  on the descriptor, pooling region
for extracting the descriptor changes as a different orientation is provided.

We therefore  use a numerical approximation  of the gradients: when  we form the
training data  we also compute  the descriptors for many  possible orientations,
every 5 degrees  in our current implementation. We can  then efficiently compute
the  derivatives   in  $\frac{\partial  \bg_{*,i}}{\partial   \theta_{*,i}}$  by
numerical  differentiation.   Note that  in  case  of  descriptors that  can  be
expressed     in    analytic     form,     for     example    learning     based
descriptors~\cite{Simo-Serra15},     we     can      also     easily     compute
$\frac{\partial \bg_{*,i}}{\partial  \theta_{*,i}}$, instead of  using numerical
approximations.

To implement the CNN $\hat{f}_\bW(.)$, we  use three convolution layers with the
ReLU activation function, each followed by  a max-pooling layer, followed by two
fully connected layers with GHH activation.  We detail the GHH activation below.
We    also   use    dropout   regularization~\cite{Srivastava14}    for   better
generalization. Implementation details are provided in Section~\ref{sec:setup}.

\subsection{Generalized Hinging Hyperplane Activation}
\label{s:ghh}

To  achieve  state-of-the-art  results  with  CNNs, we  propose  to  use  a  new
activation function in our network layers that works better for our problem than
standard ones.
This   activation  function   is   a  generalization   of   the  popular   ReLU,
maxout~\cite{Goodfellow13},   and   the  recent   PReLU~\cite{He15}   activation
functions based  on Generalized  Hinging Hyperplanes~(GHH),  which is  a general
form  for   continuous  piece-wise  linear  functions~\cite{Wang05d}.    As  GHH
activation function  is more  general, it  has less  restrictions in  shape, and
allows for more  flexibility in what a single layer  can learn.  This activation
function  plays  one  of  the  key  roles  in  our  method  for  obtaining  good
orientations, as we will show in Section~\ref{sec:GHHevaluation}.

Mathematically,         for         a         given         layer         output
$\by  =  [\by_{1,1},  \by_{1,2}, \ldots,  \by_{2,1},  \ldots,\by_{S,M}]$  before
activation, we consider the following activation function:
\begin{equation}
  o(\by) = \sum_{s\in\left\{1,2,...,S\right\}}\delta_{s}\max_{m\in\left\{1,2,...,M\right\}}\by_{s,m}
  \;\;,
  \label{eq:ghh}
\end{equation}
where 
\begin{equation}
  \delta_{s} = 
  \begin{cases}
    \hphantom{-}1, & \text{if $s$ is odd} \\
    -1, & \text{otherwise}
  \end{cases}
  \;\;,
\end{equation}
and $S$  and $M$ are meta-parameters  controlling the number of  planar segments
and thus the complexity of the function.  When $S=1$, Eq.~\eqref{eq:ghh} reduces
to  maxout  activation~\cite{Goodfellow13},  and when  additionally  $M=2$  with
$\by_{s,1}=0$, the equation  reduces to the ReLU  activation function.  Finally,
when $S=2$, $M=2$, $\by_{s,1}=0$ and  $\by_{1,m} = -\alpha_{m} \by_{2,m}$, where
$\alpha_{m}$ is a scalar variable, Eq.~\eqref{eq:ghh} is equivalent to the PReLU
activation function proposed in~\cite{He15}.

Therefore, instead of having to choose a non-linear activation function, we also
learn it  under the constraint  that it is  piece-wise linear.


\section{Results}

In this  section, we first introduce  the datasets used for  evaluation and the
setup for training our regressor. We  then demonstrate the effectiveness of our
method  by comparing  the descriptor  performances using  the original  and our
learned  orientations.  We  show  that  the best  matching  performance can  be
achieved  with our  learned orientations,  outperforming state-of-the-art.   We
also  demonstrate the  performance gain  obtained by  using the  GHH activation
compared to other  activation functions, and investigate  influence of datasets
on  the descriptor  performances.  We  finally show  a Multi-View  Stereo~(MVS)
application.\footnote{Datasets     and     source     code     available     at
  \url{http://cvlab.epfl.ch/}.}


\begin{figure*}
  \centering
  \includegraphics[width = 0.85\textwidth]{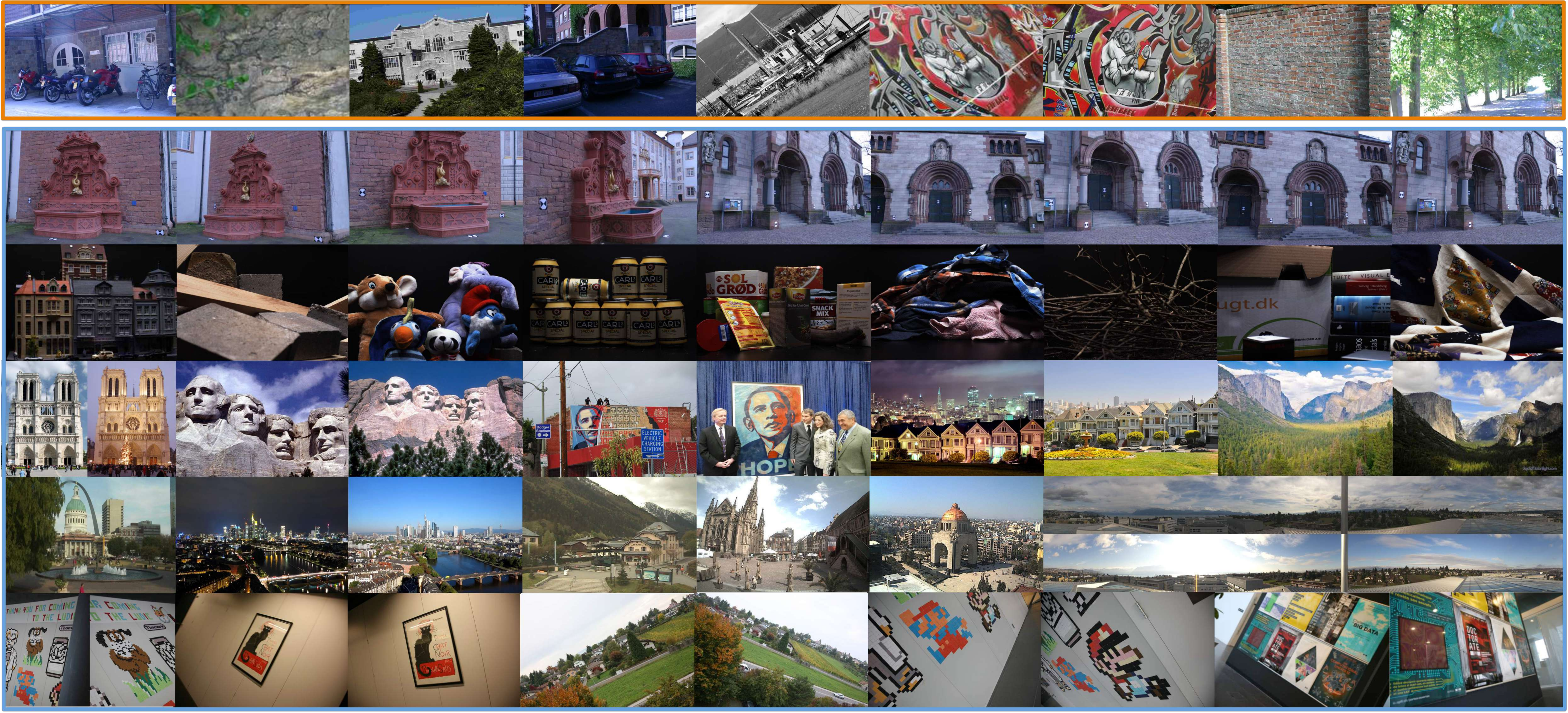}
  \caption{Selected images  from all  the datasets.  \textbf{First  row:} images
    from the  {\it Oxford}  dataset, \textbf{Second row:}  images from  the {\it
      Strecha} dataset, \textbf{Third  row:} images from the  {\it DTU} dataset,
    \textbf{Fourth row:} images  from the {\it EF}  dataset, \textbf{Fifth row:}
    images from the {\it Webcam} dataset, and \textbf{Last row:} images from our
    {\it Viewpoints} dataset.  We use the  {\it Oxford} dataset for training and
    the other datasets for testing.}
  \label{fig:dataset}
\end{figure*}


\subsection{Dataset, Training, and Evaluation  Setup}
\label{sec:setup}

\paragraph{Dataset.}

\fig{dataset} shows example images from the  datasets we use for evaluation and
training.  Note  that our  collection of  data is not  only composed  of planar
objects  but also  of 3D  objects with  self occlusions.  We also  have various
imaging changes including changes in the  camera pose.  We use the {\it Oxford}
dataset~\cite{Miko04b}   for   training,  and   the   Edge   Foci  ({\it   EF})
dataset~\cite{Zitnick11},  the {\it  Webcam} dataset~\cite{Verdie15},  the {\it
  Strecha}  dataset~\cite{Strecha08b}, the  {\it DTU}  dataset~\cite{Aanaes12},
and our own  {\it Viewpoints} dataset for testing. Details  on the datasets are
as follows:
\begin{itemize}

\item {\it Oxford} dataset~\cite{Miko04b}: 8 sequences with 48 images in total.
  The dataset  contains various imaging changes  including viewpoint, rotation,
  blur,  illumination,  scale, JPEG  compression  changes.   We thus  use  this
  dataset for training.

\item {\it  EF} dataset~\cite{Zitnick11}: 5  sequences with 38 images  in total.
  The dataset exhibits  drastic lighting changes as well as  daytime changes and
  viewpoint changes.

\item  {\it Webcam}  dataset~\cite{Verdie15}:  6 sequences  with  120 images  in
  total. The  dataset exhibits seasonal  changes as  well as daytime  changes of
  scenes taken from far away.

\item  {\it  Strecha}  dataset~\cite{Strecha08b}:  composed  of  two  sequences,
  fountain-P11 (11 images) and Herz-Jesu-P8  (8 images). The scene is non-planar
  and 3D.  The dataset exhibits large viewpoint changes with self occlusions.

\item {\it DTU} dataset~\cite{Aanaes12}: 60  sequences with 600 images in total.
  This dataset also has multiple  lighting settings for selected viewpoints, but
  we consider here only one lighting setting  as we are mostly interested in the
  changes that occur on non-planar  scenes undergoing camera movements.  We also
  sample the viewpoints  from the original dataset in regular  intervals to make
  the dataset a manageable size.

\item {\it Viewpoints} dataset: 5 sequences with 30 images in total.  We created
  our own  dataset to further  enrich the  dataset.  The dataset  exhibits large
  viewpoint changes and  in-plane rotations up to 45 degrees  from the reference
  image, which  is when commercial  cameras compensate the image  orientation as
  landscape or portrait.

\end{itemize}

\vspace{-0.8em}
\paragraph{Implementation details and training.}

We use a  patch size of $28\times28$  as input to the CNN.   For the convolution
layers, the  first convolution layer uses  a filter size of  $5\times5$ and $10$
output channels,  the second convolution layer  a filter size of  $5\times5$ and
$20$  output  channels,  and  the  third convolution  layer  a  filter  size  of
$3\times3$ and $50$ output channels.   All max-pooling layers perform $2\times2$
max pooling. The size  of the output of the first fully  connected layer is 100,
with the second  fully connected layer having two outputs  with the $\arctantwo$
mapping into orientations as described on Section~\ref{sec:formalization}.

For optimization, We use the ADAM~\cite{Kingma15} method with default parameters
and exponentially decaying learning rate. We run $100$ epochs with batch size of
$10$.   The learning  rate decay  is set  to half  the learning  rate every  ten
epochs.

Our  method is  also computationally  efficient as  we are  only estimating  the
orientations.   On  an   Intel  Xeon  E5-2680  2.5GHz   Processor,  our  current
implementation  in   Python  with   Theano~\cite{Bastien12}  takes   {\it  0.47}
milliseconds   per   feature  point   to   compute   orientations  without   any
multi-threading.  When used  with {\it SIFT} descriptors, it  overall takes {\it
  1.39} milliseconds  per feature  point to obtain  the final  descriptor.  Note
that the  C++ implementation of  the {\it MROGH}  descriptor, which is  the best
performing rotation  invariant descriptor in  our experiments, takes  {\it 1.94}
milliseconds per feature point.

\begin{figure*}
  \centering
  \includegraphics[width = 0.99\textwidth, trim=70 5 30 50,clip]{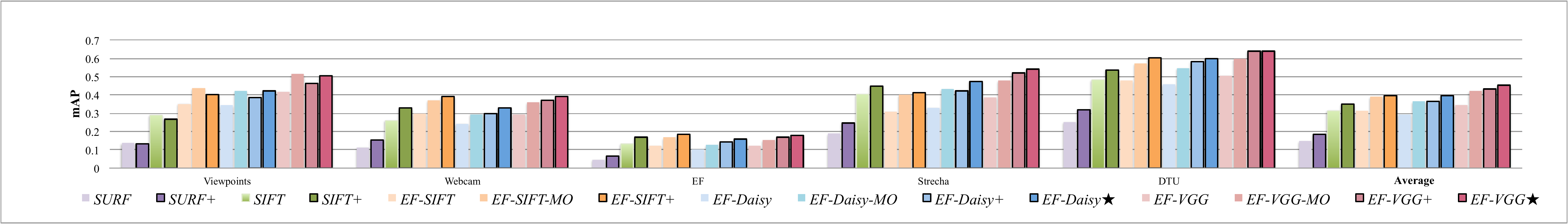}
  \caption{Descriptor performances with and without our orientation assignment.
    Methods with  the multiple  orientation strategy are  denoted with  MO, our
    orientation  assignments learned  with respective  descriptors are  denoted
    with a {\bf  +} at the end,  and those learned with {\it  SIFT} are denoted
    with  a  $\star$.  Note  that  on  average,  methods with  our  orientation
    assignments  perform better  for  all descriptors,  than using  orientation
    assignments from respective detectors and using multiple orientations.}
  \label{fig:results_withwithout}
  \vspace{-0.5em}
\end{figure*}

\vspace{-0.8em}
\paragraph{Evaluation methodology.}

To  demonstrate the  effectiveness of  our  method, we  compare the  descriptor
matching   performances  with   our   orientation   estimation  against   other
state-of-the-art descriptors\footnote{Details  on the implementations  of these
  methods are provided as appendix in the supplementary material.}.  We use the
standard  precision-recall  measure  of~\cite{Miko04b}  with  nearest  neighbor
matching, and with a maximum of 1000 feature points per image.
In case of the {\it DTU} and  {\it Strecha} datasets, the scenes are non-planar
and we rely on the 3D models and camera projection matrices to map a point from
one viewpoint  to another.   Such mapping  is used  instead of  the homography,
followed by  the overlap test  in~\cite{Miko04b}.  Results are  summarized with
the  mean  Average  Precision~(mAP)   as  in~\cite{Simonyan14},  where  mAP  is
effectively the Area Under Curve of the precision-recall graph.

We compare against  both descriptors that require  orientation estimations ({\it
  ORB}~\cite{Rublee11},       {\it       BRISK}~\cite{Leutenegger11},       {\it
  FREAK}~\cite{Alahi12}, {\it SURF}~\cite{Bay06}, {\it SIFT}~\cite{Lowe04}, {\it
  KAZE}~\cite{Alcantarilla12b},      {\it      BiCE}~\cite{Zitnick10},      {\it
  Daisy}~\cite{Tola10}, and the  learning-based {\it VGG}~\cite{Simonyan14}), as
well  as   rotation  invariant  descriptors  ({\it   LIOP}~\cite{Wang11e},  {\it
  MROGH}~\cite{Fan11},   and   {\it    sGLOH}~\cite{Bellavia10}).    Note   that
descriptors are  generally designed  for a specific  detector (for  example they
typically have different range of scale  of operation) and for fair comparisons,
we do  not interchange the detector  and descriptors.  We use  the feature point
detectors presented when  the descriptors were introduced.  In case  of the {\it
  VGG}  descriptor, we  use the  descriptor pre-learned  with the  {\it liberty}
dataset~\cite{Brown10}, as  other sequences are  partially included in  our test
set.  We employ the Edge Foci ({\it EF})~\cite{Zitnick11} detector for {\it VGG}
as it showed better performance than the Difference of Gaussians~(DoG) detector,
which was used in the original work of  {\it VGG}. We will denote this method as
{\it EF-VGG}.

We also use {\it EF} detector with the {\it Daisy} descriptor and the {\it SIFT}
descriptor, as this particular detector  was designed with these two descriptors
in mind.  We will refer them as {\it EF-Daisy} and {\it EF-SIFT}, respectively.

\subsection{Descriptor Matching Performances}
\label{s:perf}

To  demonstrate the  effectiveness of  our  method, we  evaluate the  descriptor
matching performances  with and  without our  orientation assignment.   We first
show the performance gain we obtain for {\it SIFT}, {\it SURF}, {\it Daisy}, and
{\it  VGG}   descriptors  and  then   compare  our  performance   against  other
state-of-the-art methods.
Note that  for each descriptor,  we only train our  method {\it once}  using the
{\it Oxford} dataset and test on all the other datasets.

\vspace{-0.8em}
\paragraph{Performance gain with our orientations.}
\label{s:with_without}

To  demonstrate  the  performance  gain  we obtain  by  using  our  orientation
assignments, we learned  orientations for {\it SIFT}, {\it  SURF}, {\it Daisy},
and {\it  VGG} descriptors.  We  denote descriptors computed using  our learned
orientation assignments with a {\bf +} and a $\star$ at the end; we use {\bf +}
when orientations  are learned  with respective  descriptors, and  $\star$ when
learned with  {\it SIFT}  descriptors. We also  compare against  using multiple
dominant  orientations.  Note  that  using  multiple  orientations  effectively
amounts to creating  duplicate feature points, which resulted  in 34\% increase
in descriptor  extraction time and  79\% increase  in matching time  within our
evaluation framework.

As shown in \fig{results_withwithout}, we gain a consistent boost in descriptor
matching  performance  with  our  orientation estimation.   This  includes  the
learning-based {\it  VGG} descriptor, showing that  learning-based methods also
can benefit from a better orientation assignment.  We also obtain a larger gain
on average  compared to  using multiple orientations.  The best  performance is
achieved with {\it EF-VGG}$\star$.

Interestingly, for {\it EF-Daisy} and {\it EF-VGG}, learning with the {\it SIFT}
descriptor gave larger  boost in performances than learning  with the respective
descriptors.  We  suspect that  this is  due to the  characteristics of  the two
descriptors being  less sensitive to  orientations than {\it  SIFT} descriptors,
resulting in the Jacobians with respect to orientations to vanish.

Based on the comparison results  in \fig{results_withwithout}, in the remainder
of the  results section we will  report the performance of  the best performing
handcrafted descriptor  with our  orientations, {\it EF-Daisy}$\star$,  and the
best  performing   learning-based  descriptor   with  our   orientations,  {\it
  EF-VGG}$\star$.  In Section~\ref{sec:GHHevaluation},  as the best performance
was achieved by learning with  the {\it SIFT} descriptor ({\it EF-VGG}$\star$),
we  will use  {\it  EF-SIFT}{\bf  +} to  evaluate  the  influence of  different
activation functions.

\def \resultWidth {0.86}
\begin{figure*}
  \centering
  \subfigure{
    \includegraphics[width = \resultWidth\textwidth,trim=5 25 5 50, clip]{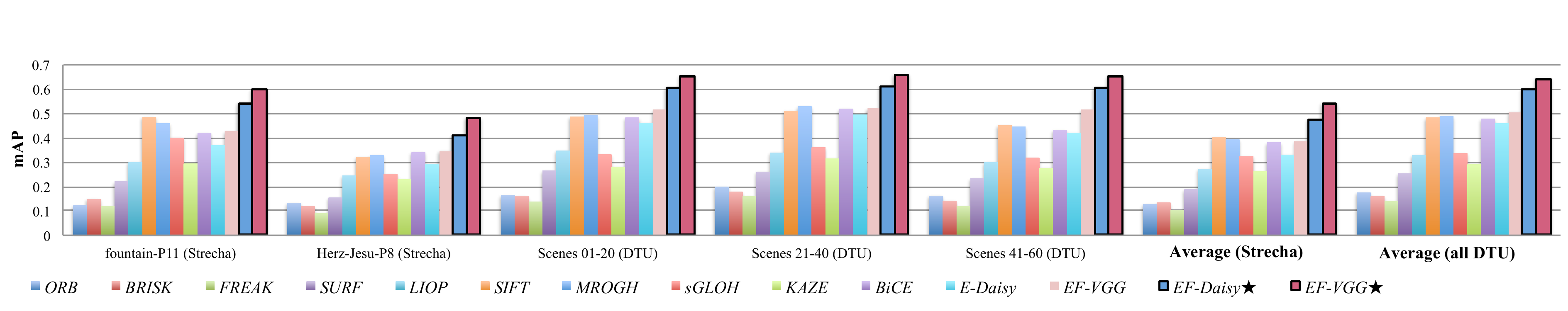}
  }
\vspace{-1.05em}

  \subfigure{
    \includegraphics[width = \resultWidth\textwidth,trim=5 30 5 40, clip]{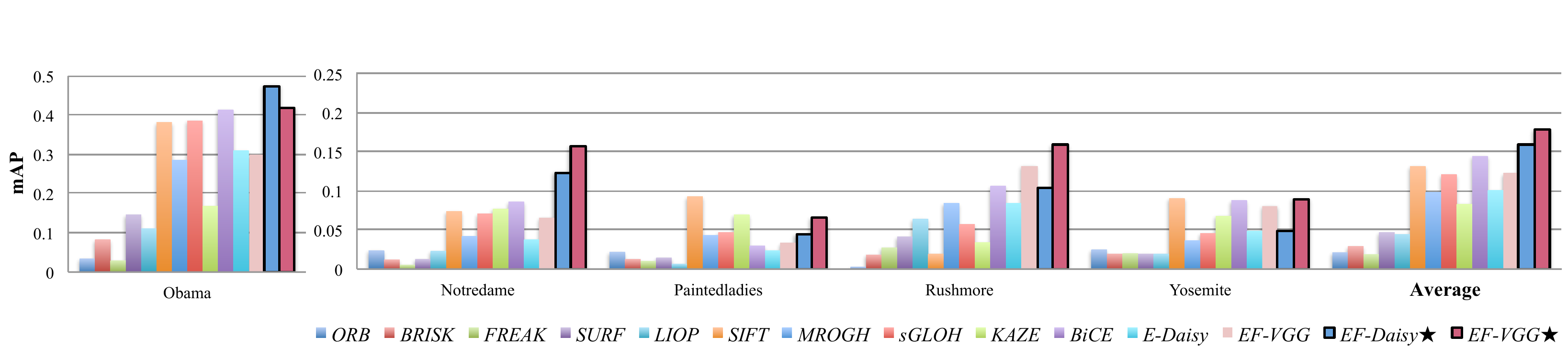}
  }
\vspace{-1.1em}

  \subfigure{
    \includegraphics[width = \resultWidth\textwidth,trim=5 25 5 50, clip]{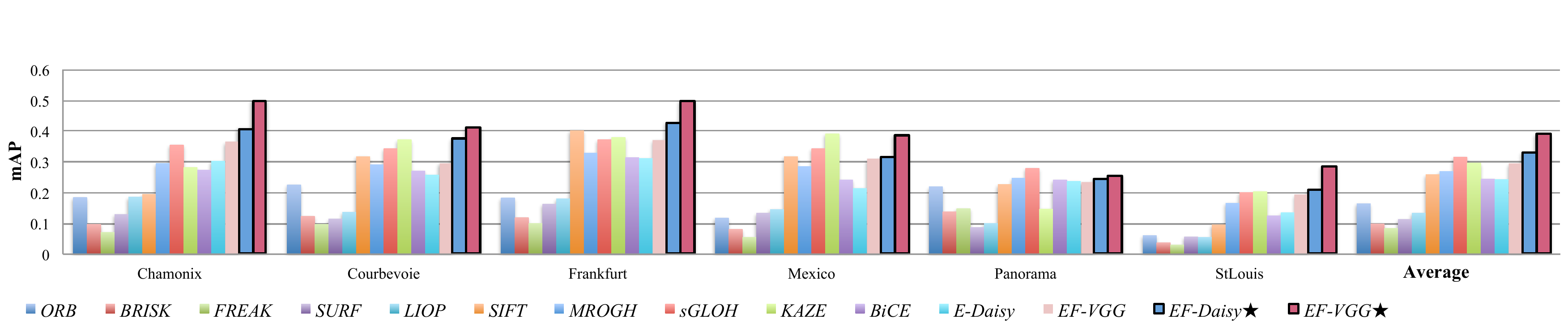}
  }
\vspace{-1.05em}

  \subfigure{
    \includegraphics[width = \resultWidth\textwidth,trim=5 5 5 50, clip]{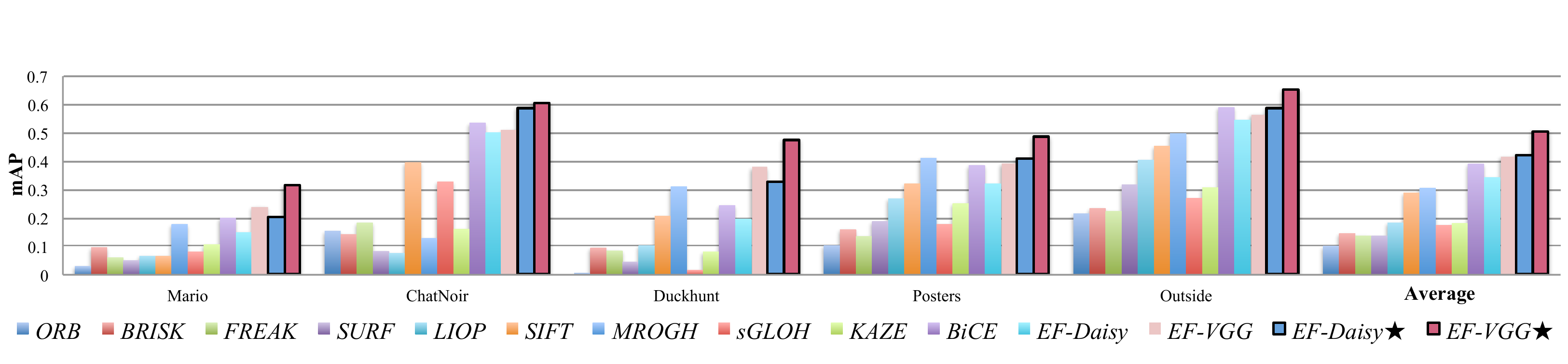}
  }

  \caption{Mean Average  Precision (mAP) for all  datasets.  \textbf{First row:}
    results  for  the   {\it  Strecha}  dataset  and  the   {\it  DTU}  dataset,
    \textbf{Second row: } results for  the {\it EF} dataset, \textbf{Third row:}
    results for the {\it Webcam} dataset  and \textbf{Last row:} results for the
    {\it Viewpoints}  dataset.  Best results  are achieved with  our orientation
    assignments, {\it EF-VGG}$\star$.}
  \label{fig:results}
\end{figure*}

\begin{table}[]

  \footnotesize
  \setlength\tabcolsep{1.5pt} 
  \definecolor{LightBlue}{rgb}{0.78,0.85,0.95}
  \centering
  \newcolumntype{L}{ >{\raggedright}m{4.65em}  }
  \newcolumntype{A}{ >{\centering}m{4.5em}  }
  \newcolumntype{B}{ >{\centering}m{3.5em}  }
  \newcolumntype{C}{ >{\centering}m{2.5em}  }
  \newcolumntype{D}{ >{\centering}m{3.5em}  }
  \newcolumntype{E}{ >{\centering}m{2.5em}  }
  \newcolumntype{F}{ >{\centering}m{2.5em}  }
  \newcolumntype{G}{ >{\centering}m{4.5em}  }

  \begin{center}
    \resizebox{0.95\columnwidth}{!}{%
      \begin{tabu}{L | A B C D E F}

        \Xhline{2\arrayrulewidth}
        \rowcolor{LightBlue}   { } & \textbf{\it Viewpoints} & \textbf{\it
          Webcam} & \textbf{\it EF} & \textbf{\it Strecha} & \textbf{\it DTU} & \textbf{Rank} \\
        \Xhline{1\arrayrulewidth}
        {\it ORB}          &13.20 & 10.33 & 11.40 & 12.50 & 12.33 & 12 \\
        {\it BRISK}        &10.40 & 12.67 & 12.60 & 12.50 & 12.43 & 13 \\
        {\it FREAK}        &11.20 & 13.33 & 12.60 & 14.00 & 13.45 & 14 \\
        {\it SURF}         &11.40 & 12.17 & 10.80 & 11.00 & 10.57 & 11 \\
        {\it LIOP}         &9.80  & 11.17 & 11.40 & 9.00  & 9.12  & 10 \\
        {\it SIFT}         &7.00  & 6.33  & 4.80  & 4.50  & 5.33  & 5  \\
        {\it MROGH}        &5.80  & 6.00  & 7.20  & 4.50  & 4.85  & 6  \\
        {\it sGLOH}        &10.20 & 3.50  & 6.00  & 7.50  & 8.93  & 8  \\
        {\it KAZE}         &9.20  & 4.83  & 6.00  & 10.00 & 9.43  & 9  \\
        {\it BiCE}         &3.80  & 7.50  & 4.00  & 5.00  & 5.28  & 4  \\
        {\it EF-Daisy}     &6.00  & 7.50  & 7.20  & 7.50  & 6.17  & 7  \\
        {\it EF-VGG}       &3.20  & 5.50  & 5.40  & 4.00  & 3.87  & 3  \\
        \Xhline{1\arrayrulewidth}
        \textbf{{\it EF-Daisy}$\star$}    &2.80  & 2.83  & 3.80  & 2.00  & 2.12  & 2  \\
        \textbf{{\it EF-VGG}$\star$}      &{\bf 1.00 }&{\bf 1.33
        }&{\bf 1.80 }&{\bf 1.00 }&{\bf 1.12 }&{\bf 1 }\\
        \Xhline{2\arrayrulewidth}
      \end{tabu}
      \hspace{0.1em}
      \begin{tabu}{G}


        \Xhline{2\arrayrulewidth}
        \rowcolor{LightBlue}    \textbf{Avg. mAP} \\
        \Xhline{1\arrayrulewidth}
        0.12\\          
        0.11\\
        0.10\\
        0.15\\
        0.19\\
        0.31\\
        0.31\\
        0.25\\
        0.22\\
        0.33\\     
        0.30\\
        0.35\\
        \Xhline{1\arrayrulewidth}
        0.40  \\
        \textbf{0.45}   \\
        \Xhline{2\arrayrulewidth}
      \end{tabu}
    }
  \end{center}
  \caption{Average rank of each method which summarizes the results in
    \fig{results}. Average rank for each dataset is
    given on the left, and the rank by averaging this value is  provided on the
    right, as well as the mAP for all datasets. Bold denotes best performance. Our method {\it EF-VGG}$\star$ ranks first, followed by {\it EF-Daisy}$\star$.}
  \label{tbl:title} 
\end{table}

\vspace{-0.8em}
\paragraph{Comparison with the state-of-the-art.}

As  shown  in  Fig.~\ref{fig:results},   both  {\it  EF-Daisy}$\star$  and  {\it
  EF-VGG}$\star$  outperform  all  compared methods,  with  {\it  EF-VGG}$\star$
outperforming all others  by a large margin.   Specifically, {\it EF-VGG}$\star$
performs {\it 27.4\%} better in terms of  mAP compared to {\it EF-VGG}, which is
the best performing  competitor.  Note that without  our orientation estimation,
although the best among the competitors, the gap is small.  Also, as pointed out
in   descriptor    performance   surveys~\cite{Aanaes12,    Khan15,   Moreels06,
  Mukherjee15}, {\it SIFT} or {\it EF-SIFT} generally give comparable results to
the state-of-the-art.

As average  results can  be influenced  by certain sequences  being too  easy or
hard, we also investigate the average rank  of each method on the entire dataset
similarly  to~\cite{Mukherjee15}.  In  \tbl{title},  we show  the  rank of  each
method on  the datasets  according to  the average  ranks of  their mAP  on each
sequence.   We also  show the  average mAP  with all  datasets for  each method.
Again, best results are obtained with our methods, {\it EF-Daisy}$\star$ and {\it EF-VGG}$\star$.

\subsection{Performance With Different Activations}
\label{sec:GHHevaluation}

To evaluate the influence of the  proposed GHH activation function, we compared
the  matching performance  of {\it  EF-SIFT}{\bf +}  with different  activation
functions. All  parameters were set to  be identical except for  the activation
type and the number of outputs in the fully connected layers.  Specifically, we
used  1600  hidden  nodes  for  ReLU,  Tanh,  and  PReLU~\cite{He15},  400  for
maxout~\cite{Goodfellow13} with  four outputs inside  the max. Note  that PReLU
has slightly more parameters than other activations, as an additional parameter
is introduced for each output of the layer.

\begin{figure}
  \centering
  \includegraphics[width = 0.42\textwidth,trim={2 5 2 45},clip]{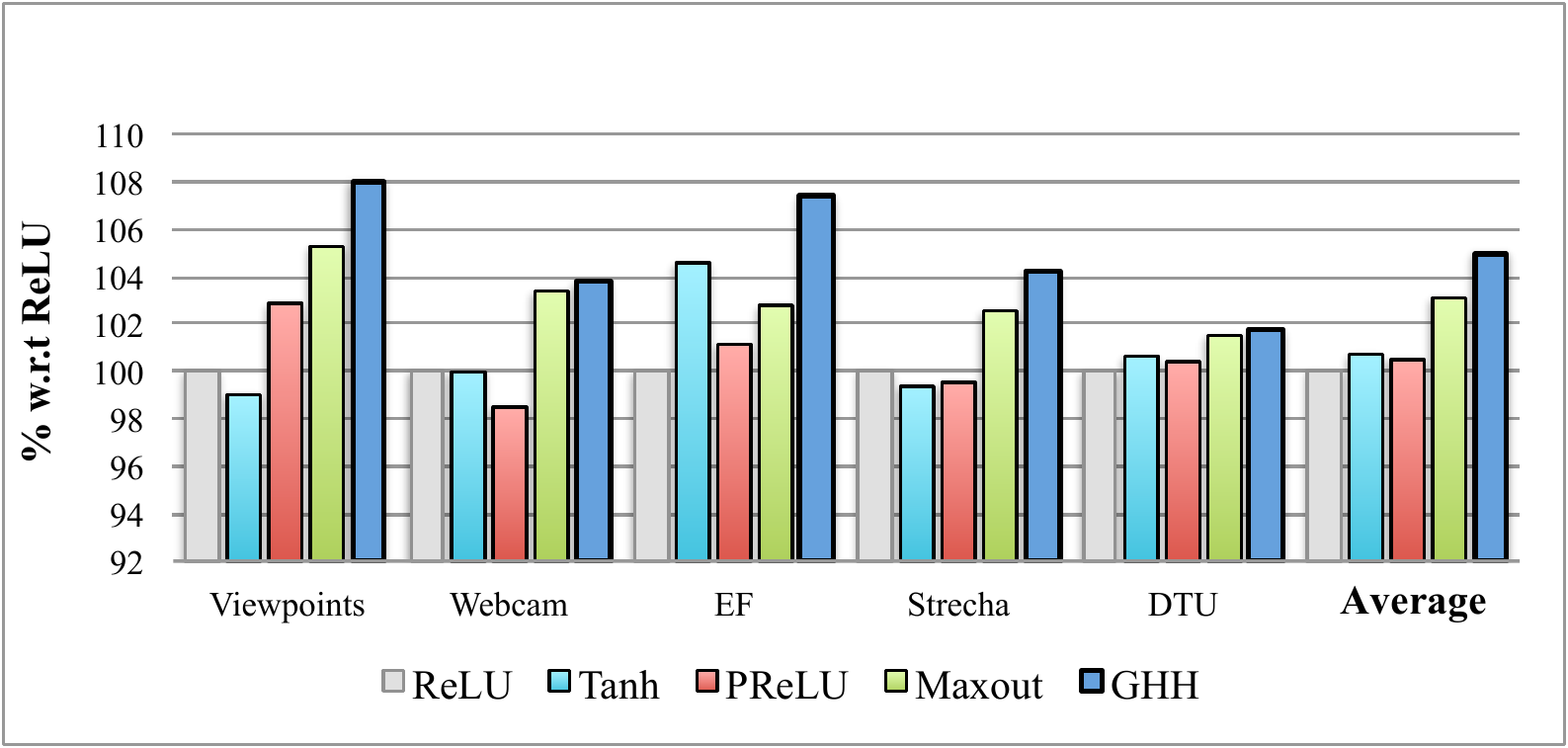}
  \caption{Relative  descriptor  performance  obtained with  the  proposed  GHH
    activation   compared  to   ReLU,   Tanh,  PReLU   and  maxout   activation
    functions. We use the mAP of ReLU activation as the reference (100\%). Best
    performance is achieved with GHH activation.}
  \label{fig:results_activation}
  \vspace{-0.5em}
\end{figure}

As shown in \fig{results_activation}, we  have a consistent gain in performance
when using the proposed GHH activation  function instead of ReLU, Tanh, maxout,
and PReLU.
This shows that  indeed using the GHH activation, which  is a generalization of
several common activation functions, is suitable for learning orientations.

\subsection{Results on Upright and Non-upright Datasets}

We observed  in the existing  datasets a general tendency  for the images  to be
carefully taken with an upright posture. As a result, it is possible to assign a
ground truth orientation to them by using a constant orientation. We will denote
this  upright assignment  of orientations  with the  suffix ``Up'',  and compare
their assignments with our orientation  assignments.  We group the datasets into
``upright''  and  ``non-upright''  ones,  depending on  whether  the  systematic
assignment to  an upright  orientation performs better  than using  the original
orientation assignments,  and compare  the performance  of {\it  EF-Daisy}, {\it
  EF-VGG}, {\it  EF-Daisy}$\star$, {\it  EF-VGG}$\star$, {\it  EF-Daisy-Up}, and
{\it EF-VGG-Up}.

\begin{figure}
  \centering
  \includegraphics[width = 0.35\textwidth,trim=5 5 5 5 ,clip]{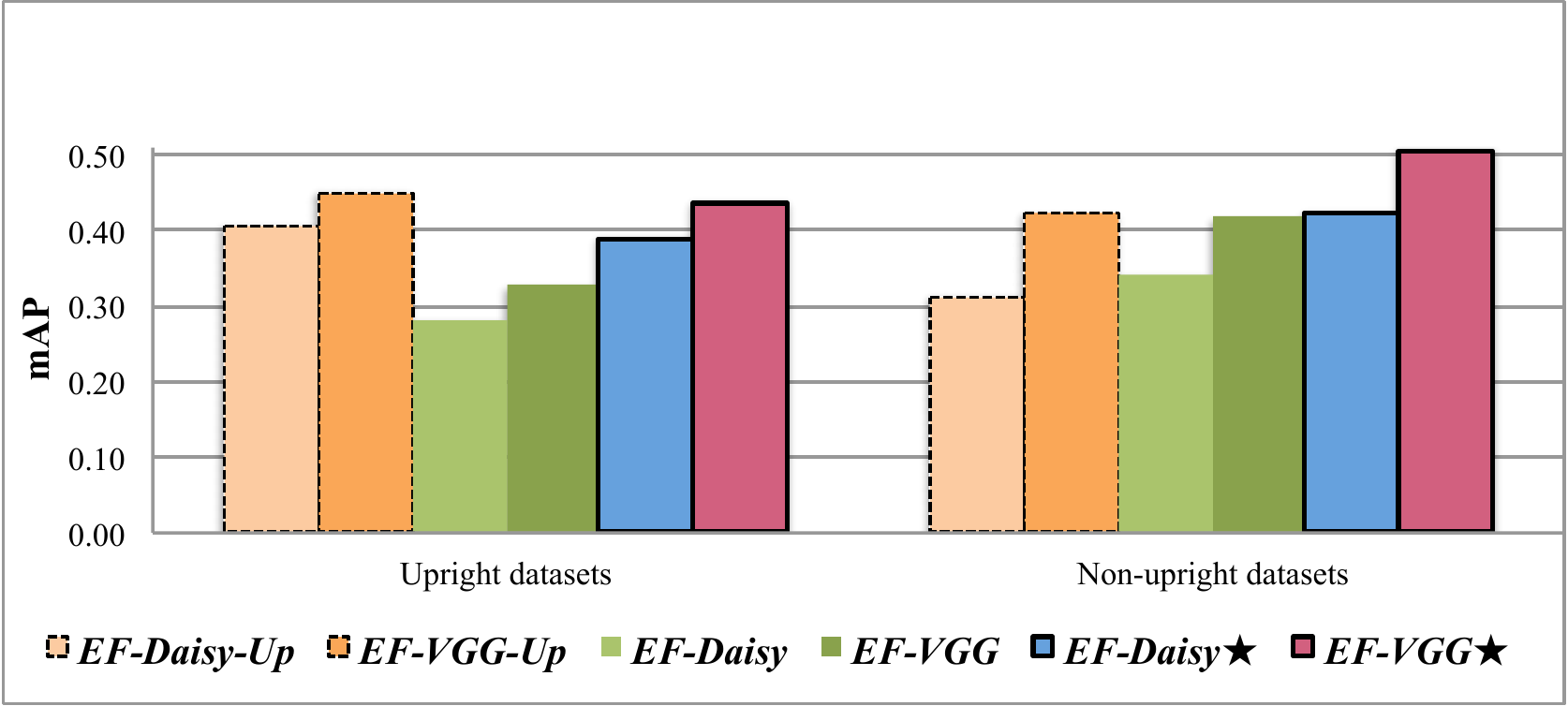}
  \caption{Performance of  {\it EF-Daisy-Up},  {\it EF-VGG-Up},  {\it EF-Daisy},
    {\it  EF-VGG}, {\it  EF-Daisy}$\star$,  and {\it  EF-VGG}$\star$ on  upright
    datasets and  non-upright datasets.  The  ``Up'' suffix in the  method names
    denotes  that  the feature  points  are  assigned zero  orientations.   {\it
      EF-Daisy}$\star$ and {\it  EF-VGG}$\star$ perform well on  both cases, and
    are  close  to  {\it  EF-Daisy-Up},  and  {\it  EF-VGG-Up}  on  the  upright
    datasets.}
  \label{fig:results_upright}
  \vspace{-0.5em}
\end{figure}

\fig{results_upright} shows  the results of  these experiments. As  expected, in
case of  upright datasets, using  a systematic upright orientation  performs the
best, which can be  seen as a upper bound for  the descriptor performances.  The
performances however degrade  when tested on non-upright  datasets. However, our
methods {\it EF-Daisy}$\star$ and {\it  EF-VGG}$\star$ perform comparably to the
upper bounds  for upright datasets  and are significantly better  on non-upright
datasets, achieving  state-of-the-art.  Note that {\it  EF-VGG} performs similar
to  {\it EF-VGG-Up},  showing that  inaccurate orientation  assignments are  not
helpful.

\subsection{Application to Multi-View Stereo}

\def \mvswidth {0.135}
\begin{figure}
  \centering

  \subfigure[{\it EF-Daisy} results]{
    \includegraphics[width = \mvswidth\textwidth,trim=340 0 540 10 ,clip]{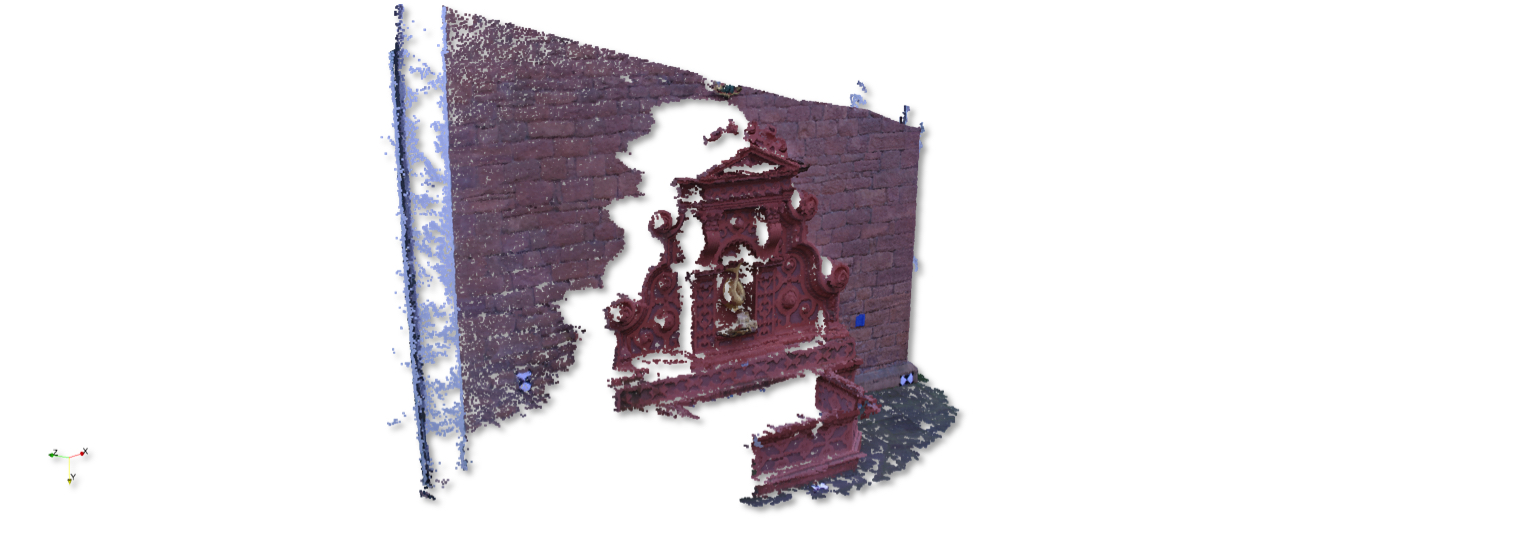}
  }
  \subfigure[{\it EF-VGG} results]{
    \includegraphics[width = \mvswidth\textwidth,trim=300 40 500 10 ,clip]{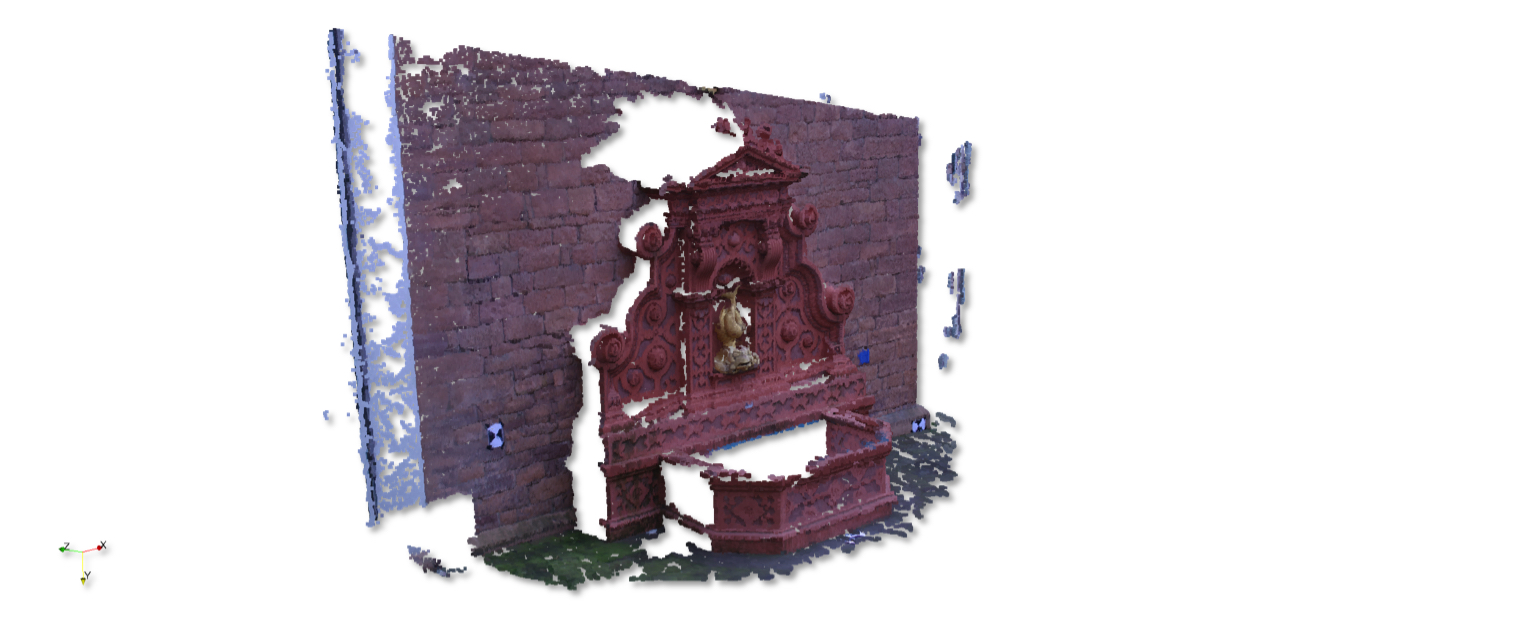}
  }
  \subfigure[{\it EF-VGG}$\star$ results]{
    \includegraphics[width = \mvswidth\textwidth,trim=300 60 500 10 ,clip]{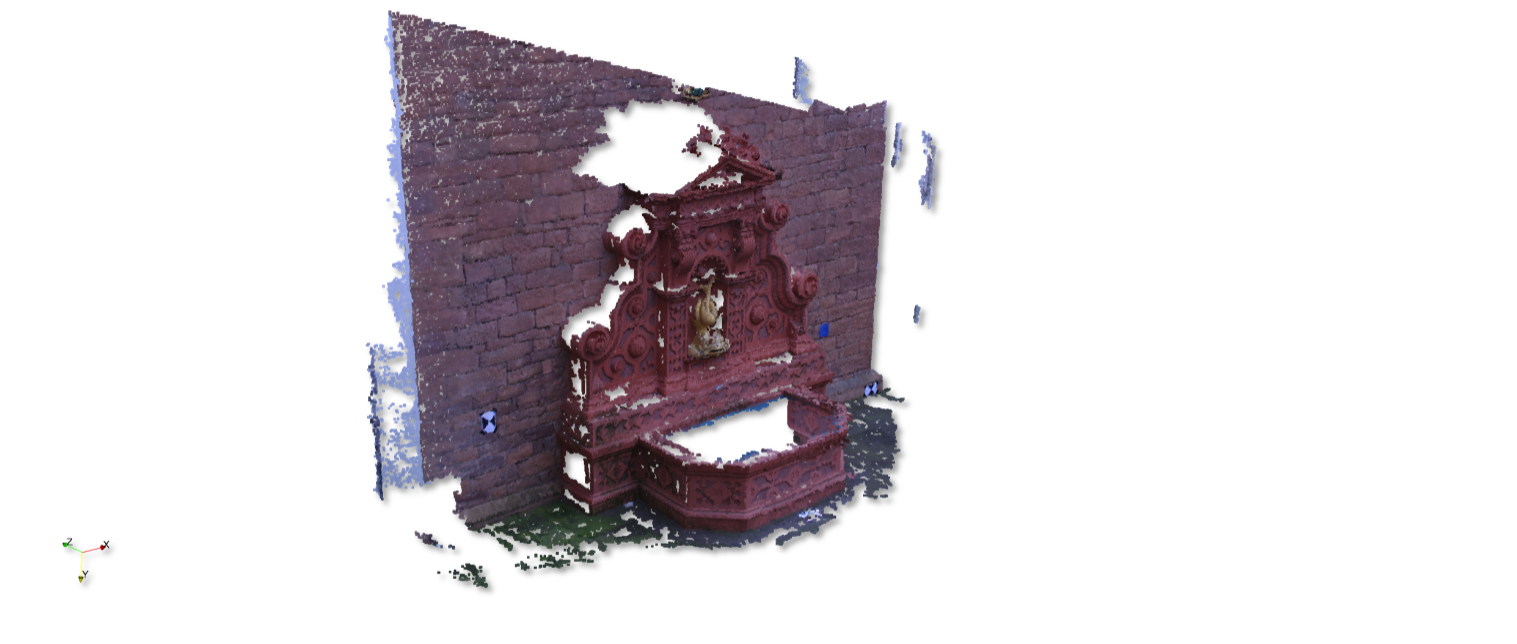}
  }

  \subfigure[{\it EF-Daisy} results]{
    \includegraphics[width = \mvswidth\textwidth,trim=340 40 360 10 ,clip]{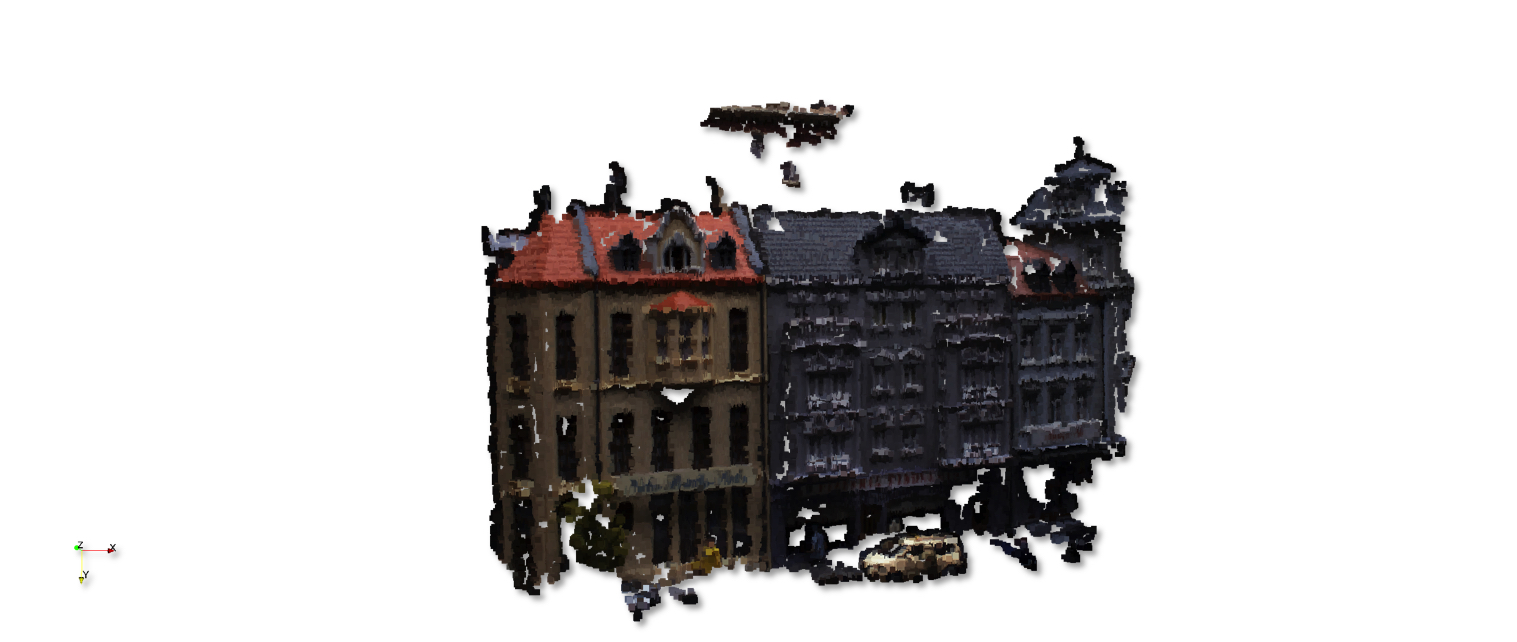}
  }
  \subfigure[{\it EF-VGG} results]{
    \includegraphics[width = \mvswidth\textwidth,trim=260 0 300 20 ,clip]{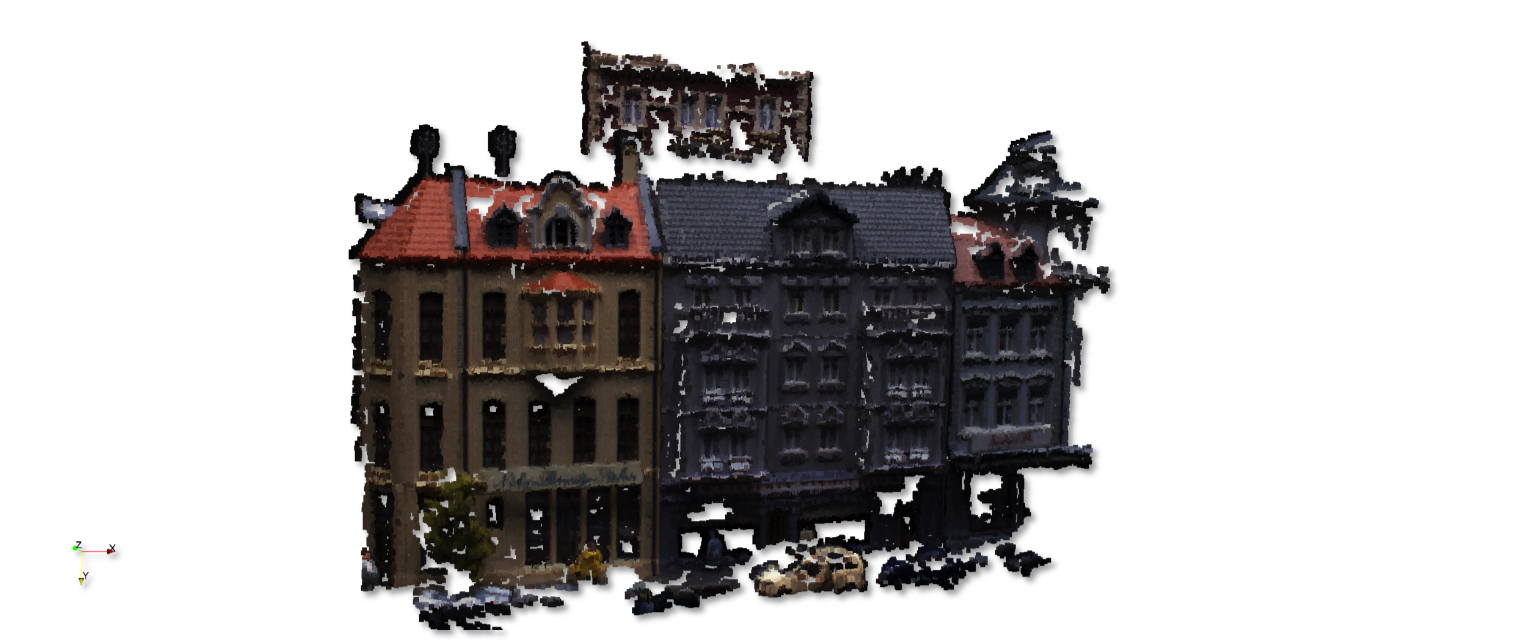}
  }
  \subfigure[{\it EF-VGG}$\star$ results]{
    \includegraphics[width = \mvswidth\textwidth,trim=350 20 350 10 ,clip]{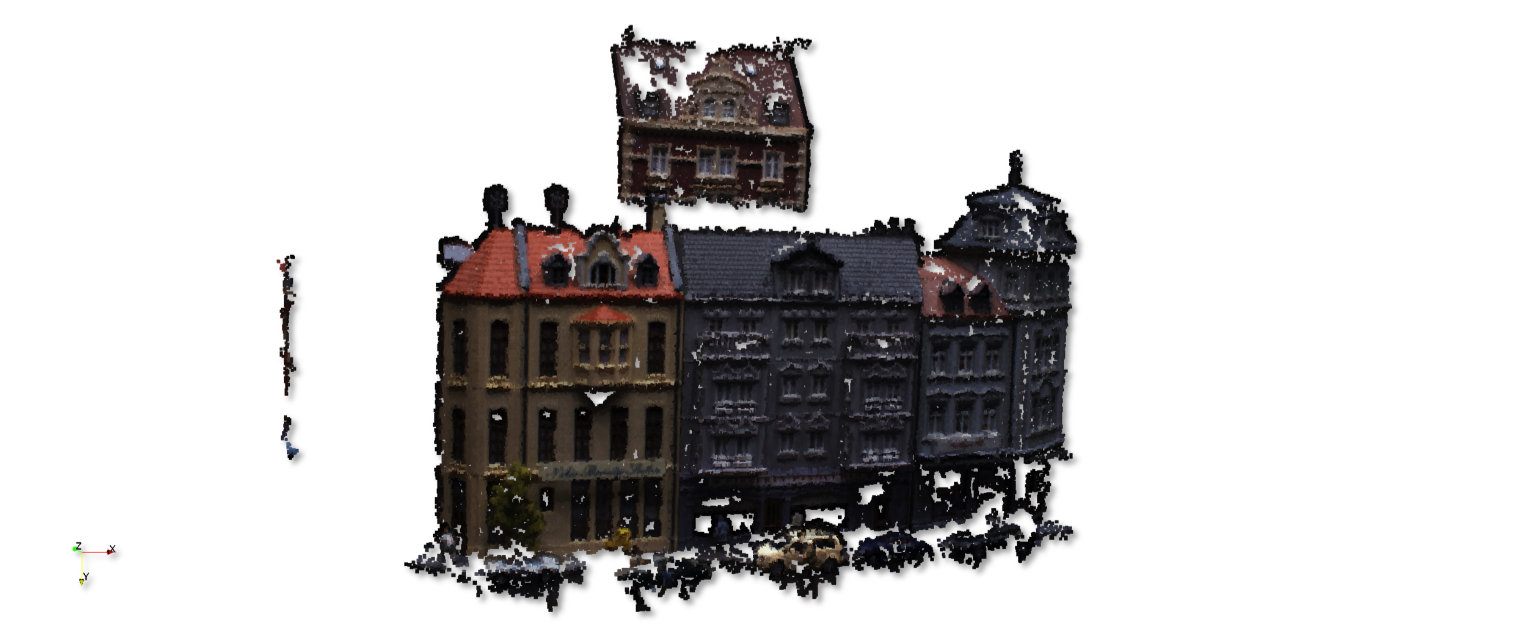}
  }



  \caption{Multi-View  Stereo  (MVS) application~\cite{Wu13,Wu11}  example  with
    {\it   EF-Daisy},  {\it   EF-VGG}  and   our  method   {\it  EF-VGG}$\star$.
    \mbox{(a)~--~(c)} results  for the  fountain sequence  of the  {\it Strecha}
    dataset,  \mbox{(d)~--~(f)}  results for  the  Scene  55  of the  {\it  DTU}
    dataset.  Original images  are shown as the first image  for each dataset in
    \fig{dataset}.   As better  matches are  provided with  {\it EF-VGG}$\star$,
    more detailed MVS reconstructions are obtained. Same {\it EF} feature points
    were used, differing only in orientation assignments.}
  \label{fig:mvs}
\end{figure}

We     also    apply     our     orientation    estimations     for    a     MVS
application~\cite{Wu13,Wu11}.  \fig{mvs} shows MVS results using {\it EF-Daisy},
{\it EF-VGG} and {\it EF-VGG}$\star$.   Due to better matching performances, our
method {\it  EF-VGG}$\star$ gives  best MVS results,  followed by  {\it EF-VGG}.
Specifically, for the fountain sequence of  the {\it Strecha} dataset, we obtain
$260747$ vertices with {\it EF-Daisy},  $323979$ with {\it EF-VGG}, and $365261$
with {\it  EF-VGG}$\star$.  For  Scene 55  of the {\it  DTU} dataset,  we obtain
$72972$, $86826$,  and $95014$  vertices for {\it  EF-Daisy}, {\it  EF-VGG}, and
{\it EF-VGG}$\star$, respectively.


\section{Conclusion}

We have  introduced a learning scheme  using a Convolutional Neural  Network for
the estimation of a canonical orientation for feature points, which improves the
performance of  existing descriptors.  We  proposed to train Siamese  network to
predict an orientation, which avoided the need of explicitly defining a ``good''
orientation to  learn. We  also proposed  a new  GHH activation  function, which
generalizes existing piece-wise linear  activation functions and performs better
for our  task. We  evaluated the  effectiveness of  our learned  orientations by
comparing  the   descriptor  performances  with  and   without  our  orientation
assignment.  Descriptors  using our  orientations gained  consistent performance
increase  and outperformed  state-of-the-art  descriptors on  all datasets.   We
finally investigated  the influence of  the GHH activation function  showing its
effectiveness.

Although we were able to enhance the performance of the learning-based {\it VGG}
descriptor  as  well, an  interesting  future  research  direction is  to  fully
integrate  our  method  with  learning-based descriptors,  such  as  the  recent
descriptor presented in~\cite{Simo-Serra15}.  In which case, we can have a fully
differentiable Siamese network which learns  both the orientation assignment and
the descriptor at the same time.

\section*{Acknowledgement}

This work was supported in part by the EU FP7 project MAGELLAN under the grant number ICT-FP7-611526 and in part by the EU project EDUSAFE.



{\small
\bibliographystyle{ieee}
\bibliography{short,vision,learning}

\begin{thebibliography}{10}\itemsep=-1pt

\bibitem{Aanaes12}
H.~Aan{\ae}s, A.~L. Dahl, and K.~S. Pedersen.
\newblock {Interesting Interest Points}.
\newblock {\em IJCV}, 97:18--35, 2012.

\bibitem{Alahi12}
A.~Alahi, R.~Ortiz, and P.~Vandergheynst.
\newblock {FREAK: Fast Retina Keypoint}.
\newblock In {\em CVPR}, 2012.

\bibitem{Alcantarilla12b}
P.~Alcantarilla, P.~Fern{\'a}ndez, A.~Bartoli, and A.~J. Davidson.
\newblock {KAZE Features}.
\newblock In {\em ECCV}, 2012.

\bibitem{Allaire08}
S.~Allaire, J.~J. Kim, S.~L. Breen, D.~A. Jaffray, and V.~Pekar.
\newblock {Full Orientation Invariance and Improved Feature Selectivity of 3D
  SIFT with Application to Medical Image Analysis}.
\newblock In {\em CVPR}, 2008.

\bibitem{Bastien12}
F.~Bastien, P.~Lamblin, R.~Pascanu, J.~Bergstra, I.~Goodfellow., A.~Bergeron,
  N.~Bouchard, and Y.~Bengio.
\newblock {Theano: New Features and Speed Improvements}.
\newblock In {\em NIPS}, 2012.

\bibitem{Bay06}
H.~Bay, T.~Tuytelaars, and L.~{Van~Gool}.
\newblock {{SURF}: Speeded Up Robust Features}.
\newblock In {\em ECCV}, 2006.

\bibitem{Bellavia10}
F.~Bellavia and D.~Tegolo.
\newblock {Improving Sift-Based Descriptors Stability to Rotations}.
\newblock In {\em ICPR}, 2010.

\bibitem{Bellavia14}
F.~Bellavia, D.~Tegolo, and C.~Valenti.
\newblock {Keypoint Descriptor Matching with Context-Based Orientation
  Estimation}.
\newblock {\em IVC}, 32(9):559--567, 2014.

\bibitem{Bromley93}
J.~Bromley, I.~Guyon, Y.~LeCun, E.~Sackinger, and R.~Shah.
\newblock {Signature Verification Using a Siamese Time Delay Neural Network}.
\newblock In {\em NIPS}, 1993.

\bibitem{Brown10}
M.~Brown, G.~Hua, and S.~Winder.
\newblock {Discriminative Learning of Local Image Descriptors}.
\newblock {\em PAMI}, 2011.

\bibitem{Brown05a}
M.~Brown, R.~Szeliski, and S.~Winder.
\newblock {Multi-Image Matching Using Multi-Scale Oriented Patches}.
\newblock In {\em CVPR}, 2005.

\bibitem{Chopra05}
S.~Chopra, R.~Hadsell, and Y.~LeCun.
\newblock {Learning a Similarity Metric Discriminatively, with Application to
  Face Verification}.
\newblock In {\em CVPR}, 2005.

\bibitem{Fan14}
B.~Fan, Q.~Kong, T.~Trzcinski, Z.~Wang, C.~Pan, and P.~Fua.
\newblock {Receptive Fields Selection for Binary Feature Description}.
\newblock {\em TIP}, 23(6):2583--2595, 2014.

\bibitem{Fan11}
B.~Fan, F.~Wu, and Z.~Hu.
\newblock {Aggregating Gradient Distributions into Intensity Orders: A Novel
  Local Image Descriptor}.
\newblock In {\em CVPR}, 2011.

\bibitem{Gauglitz11}
S.~Gauglitz, M.~Turk, and T.~H{\"o}llerer.
\newblock {Improving Keypoint Orientation Assignment}.
\newblock In {\em BMVC}, 2011.

\bibitem{Goodfellow13}
I.~Goodfellow, D.~Warde-farley, M.~Mirza, A.~Courville, and Y.~Bengio.
\newblock {Maxout Networks}.
\newblock In {\em ICML}, 2013.

\bibitem{He15}
K.~He, X.~Zhang, R.~Ren, and J.~Sun.
\newblock {Delving Deep into Rectifiers: Surpassing Human-Level Performance on
  Imagenet Classification}.
\newblock In {\em ICCV}, 2015.

\bibitem{Hinterstoisser08a}
S.~Hinterstoisser, S.~Benhimane, N.~Navab, P.~Fua, and V.~Lepetit.
\newblock {Online Learning of Patch Perspective Rectification for Efficient
  Object Detection}.
\newblock In {\em CVPR}, 2008.

\bibitem{Khan15}
N.~Khan, B.~Mccane, and S.~Mills.
\newblock {Better Than SIFT?}
\newblock {\em MVA}, 26(6):819--836, 2015.

\bibitem{Kingma15}
D.~Kingma and J.~Ba.
\newblock {Adam: {A} Method for Stochastic Optimisation}.
\newblock In {\em ICLR}, 2015.

\bibitem{Lazebnik04}
S.~Lazebnik, C.~Schmid, and J.~Ponce.
\newblock {Semi-Local Affine Parts for Object Recognition}.
\newblock In {\em BMVC}, 2004.

\bibitem{Leutenegger11}
S.~Leutenegger, M.~Chli, and R.~Siegwart.
\newblock {{BRISK}: Binary Robust Invariant Scalable Keypoints}.
\newblock In {\em ICCV}, 2011.

\bibitem{Lin12}
W.-Y. Lin, L.~Liu, Y.~Matsushita, K.-L. Low, and S.~Liu.
\newblock {Aligning Images in the Wild}.
\newblock In {\em CVPR}, 2012.

\bibitem{Liu14}
K.~Liu, H.~Skibbe, T.~Schmidt, T.~Blein, K.~Palme, T.~Brox, and O.~Ronneberger.
\newblock {Rotation-Invariant HOG Descriptors Using Fourier Analysis in Polar
  and Spherical Coordinates}.
\newblock {\em IJCV}, 106(3):342--364, 2014.

\bibitem{Lowe04}
D.~Lowe.
\newblock {Distinctive Image Features from Scale-Invariant Keypoints}.
\newblock {\em IJCV}, 20(2), 2004.

\bibitem{Miko04b}
K.~Mikolajczyk and C.~Schmid.
\newblock {A Performance Evaluation of Local Descriptors}.
\newblock {\em PAMI}, 27(10):1615--1630, 2004.

\bibitem{Miko04c}
K.~Mikolajczyk and C.~Schmid.
\newblock {Scale and Affine Invariant Interest Point Detectors}.
\newblock {\em IJCV}, 60:63--86, 2004.

\bibitem{Moreels06}
P.~Moreels and P.~Perona.
\newblock {Evaluation of Features Detectors and Descriptors Base on 3D
  Objects}.
\newblock In {\em IJCV}, 2006.

\bibitem{Mukherjee15}
D.~Mukherjee, Q.~M.~J. Wu, and G.~Wang.
\newblock {A Comparative Experimental Study of Image Feature Detectors and
  Descriptors}.
\newblock {\em MVA}, 26(4):443--466, 2015.

\bibitem{Osadchy07}
M.~Osadchy, Y.~LeCun, and M.~Miller.
\newblock {Synergistic Face Detection and Pose Estimation with Energy-Based
  Models}.
\newblock {\em JMLR}, 8:1197--1215, 2007.

\bibitem{Penedones11}
H.~Penedones, R.~Collobert, F.~Fleuret, and D.~Grangier.
\newblock {Improving Object Classification using Pose Information }.
\newblock Technical report, Idiap Research Institute, 2011.

\bibitem{Rublee11}
E.~Rublee, V.~Rabaud, K.~Konolidge, and G.~Bradski.
\newblock {ORB: An Efficient Alternative to SIFT or SURF}.
\newblock In {\em ICCV}, 2011.

\bibitem{Simo-Serra15}
E.~{Simo-Serra}, E.~Trulls, L.~Ferraz, I.~Kokkinos, P.~Fua, and
  F.~{Moreno-Noguer}.
\newblock {Discriminative Learning of Deep Convolutional Feature Point
  Descriptors}.
\newblock In {\em ICCV}, 2015.

\bibitem{Simonyan14}
K.~Simonyan, A.~Vedaldi, and A.~Zisserman.
\newblock {Learning Local Feature Descriptors Using Convex Optimisation}.
\newblock {\em PAMI}, 2014.

\bibitem{Srivastava14}
N.~Srivastava, G.~Hinton, A.~Krizhevsky, I.~Sutskever, and R.~Salakhutdinov.
\newblock {Dropout: A Simple Way to Prevent Neural Networks from Overfitting}.
\newblock {\em JMLR}, 15:1929--1958, 2014.

\bibitem{Strecha08b}
C.~Strecha, W.~Hansen, L.~{Van~Gool}, P.~Fua, and U.~Thoennessen.
\newblock {On Benchmarking Camera Calibration and Multi-View Stereo for High
  Resolution Imagery}.
\newblock In {\em CVPR}, 2008.

\bibitem{Taylor11}
S.~Taylor and T.~Drummond.
\newblock {Binary Histogrammed Intensity Patches for Efficient and Robust
  Matching}.
\newblock {\em IJCV}, 94(2):241--265, 2011.

\bibitem{Tola10}
E.~Tola, V.~Lepetit, and P.~Fua.
\newblock {Daisy: An Efficient Dense Descriptor Applied to Wide Baseline
  Stereo}.
\newblock {\em PAMI}, 32(5):815--830, 2010.

\bibitem{Trzcinski15}
T.~Trzcinski, M.~Christoudias, and V.~Lepetit.
\newblock {Learning Image Descriptors with Boosting}.
\newblock {\em PAMI}, 2015.

\bibitem{Verdie15}
Y.~Verdie, K.~M. Yi, P.~Fua, and V.~Lepetit.
\newblock {{TILDE}: A Temporally Invariant Learned {DEtector}}.
\newblock In {\em CVPR}, 2015.

\bibitem{Wang05d}
S.~Wang and X.~Sun.
\newblock {Generalization of Hinging Hyperplanes}.
\newblock {\em TIT}, 51(12):4425--4431, 2005.

\bibitem{Wang11e}
Z.~Wang, B.~Fan, and F.~Wu.
\newblock {Local Intensity Order Pattern for Feature Description}.
\newblock In {\em ICCV}, 2011.

\bibitem{Winder09}
S.~Winder, G.~Hua, and M.~Brown.
\newblock {Picking the Best Daisy}.
\newblock In {\em CVPR}, June 2009.

\bibitem{Wu13}
C.~Wu.
\newblock {Towards Linear-Time Incremental Structure from Motion}.
\newblock In {\em 3DV}, 2013.

\bibitem{Wu11}
C.~Wu, S.~Agarwal, B.~Curless, and S.~M. Seitz.
\newblock {Multicore Bundle Adjustment}.
\newblock In {\em CVPR}, 2011.

\bibitem{Zagoruyko15}
S.~Zagoruyko and N.~Komodakis.
\newblock {Learning to Compare Image Patches via Convolutional Neural
  Networks}.
\newblock In {\em CVPR}, 2015.

\bibitem{Zitnick10}
C.~Zitnick.
\newblock {Binary Coherent Edge Descriptors}.
\newblock In {\em ECCV}, 2010.

\bibitem{Zitnick11}
C.~Zitnick and K.~Ramnath.
\newblock {Edge Foci Interest Points}.
\newblock In {\em ICCV}, 2011.

\end{thebibliography}
}

\cleardoublepage
\onecolumn
\appendix 

\section{Supplementary Appendix}

In this appendix, we provide details on the implementations used in the experiments.

\subsection{Implementations of the Methods}

\paragraph{Compared Methods}
To keep  the maximum  number of features  points to 1000,  we sort  the detected
feature points according  to their respective response scores and  keep the best
1000.  Details for the implementations of the compared methods are as follows:

\begin{itemize}
\item {\it ORB}~\cite{Rublee11}: OpenCV library -- \url{http://opencv.org/downloads.html} \\
  We used nFeatures=1000, nLevels=3, and default values for other parameters.
\item {\it BRISK}~\cite{Leutenegger11}: Provided by the authors -- \url{http://www.asl.ethz.ch/people/lestefan/personal/BRISK} \\
  We used threshold of 20, with default values for other parameters.
\item {\it FREAK}~\cite{Alahi12}: Provided by the authors  -- \url{https://github.com/kikohs/freak} \\
  Default parameters were used.
\item {\it SURF}~\cite{Bay06}: OpenCV library -- \url{http://opencv.org/downloads.html} \\
  Default parameters were used.
\item {\it LIOP}~\cite{Wang11e}: VLFeat library -- \url{http://www.vlfeat.org/} \\
  Default parameters were used.
\item {\it SIFT}~\cite{Lowe04}: OpenCV library -- \url{http://opencv.org/downloads.html} \\
  Default parameters were used.
\item {\it MROGH}~\cite{Fan11}: Provided by the authors -- \url{https://github.com/bfan/MROGH-feature-descriptor} \\
  Default parameters were used.
\item {\it sGLOH}~\cite{Bellavia10}: Provided by the authors -- \url{http://www.math.unipa.it/fbellavia/htm/research.html} \\
  Default parameters were used.
\item {\it KAZE}~\cite{Alcantarilla12b}: Provided by the authors -- \url{https://github.com/pablofdezalc/kaze} \\
  Default parameters were used.
\item {\it EF}~\cite{Zitnick11} and {\it BiCE}~\cite{Zitnick10}: Provided by the authors -- \url{http://research.microsoft.com/en-us/um/people/larryz/edgefoci/edge_foci.htm} \\
  Default parameters were used.
\item {\it Daisy}~\cite{Zitnick10}: Provided by the authors -- \url{https://github.com/etola/libdaisy} \\
  Patches were extracted to be four times the scale, which was the value authors
  used in~\cite{Zitnick11}. Other parameters are set to default values.
\item {\it VGG}~\cite{Simonyan14}: Provided by the authors --
  \url{http://www.robots.ox.ac.uk/~vgg/software/learn_desc/} \\
  Patches  were extracted  with the  VLFeat  library, with  a relativeExtent  of
  $7.5$, which is the same as what {\it SIFT} uses. We use the pre-learned model
  learned with  the {\it liberty}  dataset from~\cite{Simonyan14}, as  the other
  two datasets are partially included in our test set.
\end{itemize}

\paragraph{Our Methods}\

We  used  the  Python  Theano  library~\cite{Bastien12}  for  implementation  --
\url{http://deeplearning.net/software/theano/}

\end{document}